\documentclass[10pt,twocolumn,letterpaper]{article}

\usepackage{cvpr}              

\usepackage{algorithm}
\usepackage{algorithmic}
\usepackage[utf8]{inputenc} 
\usepackage[T1]{fontenc}    
\usepackage{hyperref}       
\usepackage{url}            
\usepackage{booktabs}       
\usepackage{amsfonts}       
\usepackage{nicefrac}       
\usepackage{microtype}      
\usepackage{xcolor}         
\usepackage{multirow}
\usepackage{float}
\usepackage{graphicx}
\usepackage{colortbl}
\usepackage{subcaption}
\usepackage{caption}
\usepackage{tabularx}
\usepackage{physics}
\usepackage{bbding}
\usepackage{multirow}
\usepackage{booktabs}
\usepackage{appendix}
\definecolor{lightpink}{rgb}{0.9, 0.95, 1.0}  

\usepackage{amsmath,amsfonts,bm}

\def\eqref#1{equation~\ref{#1}}

\def\1{\bm{1}}

\def\vt{{\bm{t}}}

\def\vv{{\bm{v}}}

\def\vx{{\bm{x}}}

\DeclareMathAlphabet{\mathsfit}{\encodingdefault}{\sfdefault}{m}{sl}
\SetMathAlphabet{\mathsfit}{bold}{\encodingdefault}{\sfdefault}{bx}{n}

\usepackage{amsmath}
\usepackage{parcolumns}
\usepackage{picinpar}
\usepackage[export]{adjustbox}
\newcommand*\samethanks[1][\value{footnote}]{\footnotemark[#1]}

\definecolor{cvprblue}{rgb}{0.21,0.49,0.74}

\title{
    \raisebox{0.2\height}{\begin{minipage}{0.06\textwidth}
        \centering
        \includegraphics[width=\linewidth]{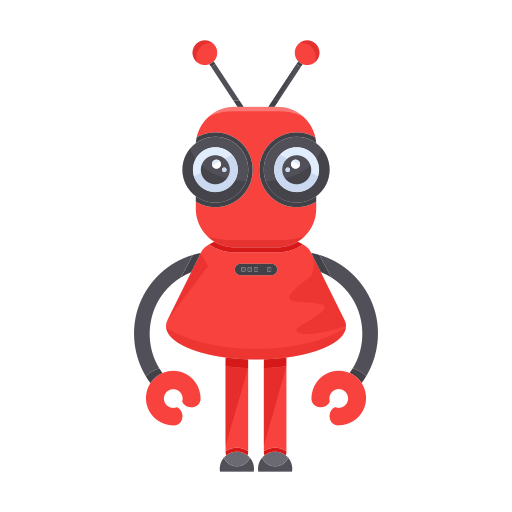}
    \end{minipage}}%
    \begin{minipage}[t]{0.84\textwidth}
        \centering
        ANTS: Adaptive Negative Textual Space Shaping for OOD Detection \\
        via Test-Time MLLM Understanding and Reasoning
    \end{minipage}
}

\author{
Wenjie Zhu$^{1,2}$\thanks{These authors contributed equally to this work.} 
\quad 
Yabin Zhang$^{3}$\samethanks 
\quad 
Xin Jin$^{2,4}$  
\quad 
Wenjun Zeng$^{2}$\thanks{Corresponding authors.} 
\quad 
Lei Zhang$^{1}$\samethanks
\\
$^1$The Hong Kong Polytechnic University
\qquad 
$^2$Eastern Institute of Technology, Ningbo  
\\
\qquad 
$^3$Harbin Institute of Technology (Shenzhen)
\qquad 
$^4$Zhongguancun Academy
\\
{\tt\small  22040319r@connect.polyu.hk,} 
{\tt\small wzeng-vp@eitech.edu.cn, cslzhang@comp.polyu.edu.hk} \\
}

\begin{document}
\maketitle
\begin{abstract}
The introduction of negative labels (NLs) has proven effective in enhancing Out-of-Distribution (OOD) detection. However, existing methods often lack an understanding of OOD images, making it difficult to construct an accurate negative space. Furthermore, the absence of negative labels semantically similar to ID labels constrains their capability in near-OOD detection. To address these issues, we propose shaping an Adaptive Negative Textual Space (ANTS) by leveraging the understanding and reasoning capabilities of multimodal large language models (MLLMs). Specifically, we cache images likely to be OOD samples from the historical test images and prompt the MLLM to describe these images, generating expressive negative sentences that precisely characterize the OOD distribution and enhance far-OOD detection. For the near-OOD setting, where OOD samples resemble the in-distribution (ID) subset, we cache the subset of ID classes that are visually similar to historical test images and then leverage MLLM reasoning to generate visually similar negative labels tailored to this subset, effectively reducing false negatives and improving near-OOD detection. To balance these two types of negative textual spaces, we design an adaptive weighted score that enables the method to handle different OOD task settings (near-OOD and far-OOD), making it highly adaptable in open environments. On the ImageNet benchmark, our ANTS significantly reduces the FPR95 by 3.1\%, establishing a new state-of-the-art. Furthermore, our method is training-free and zero-shot, enabling high scalability. Codes are available at \url{https://github.com/ZhuWenjie98/ANTS}.
\end{abstract}

\begin{figure}[h]
    \centering
    \includegraphics[width=\linewidth]{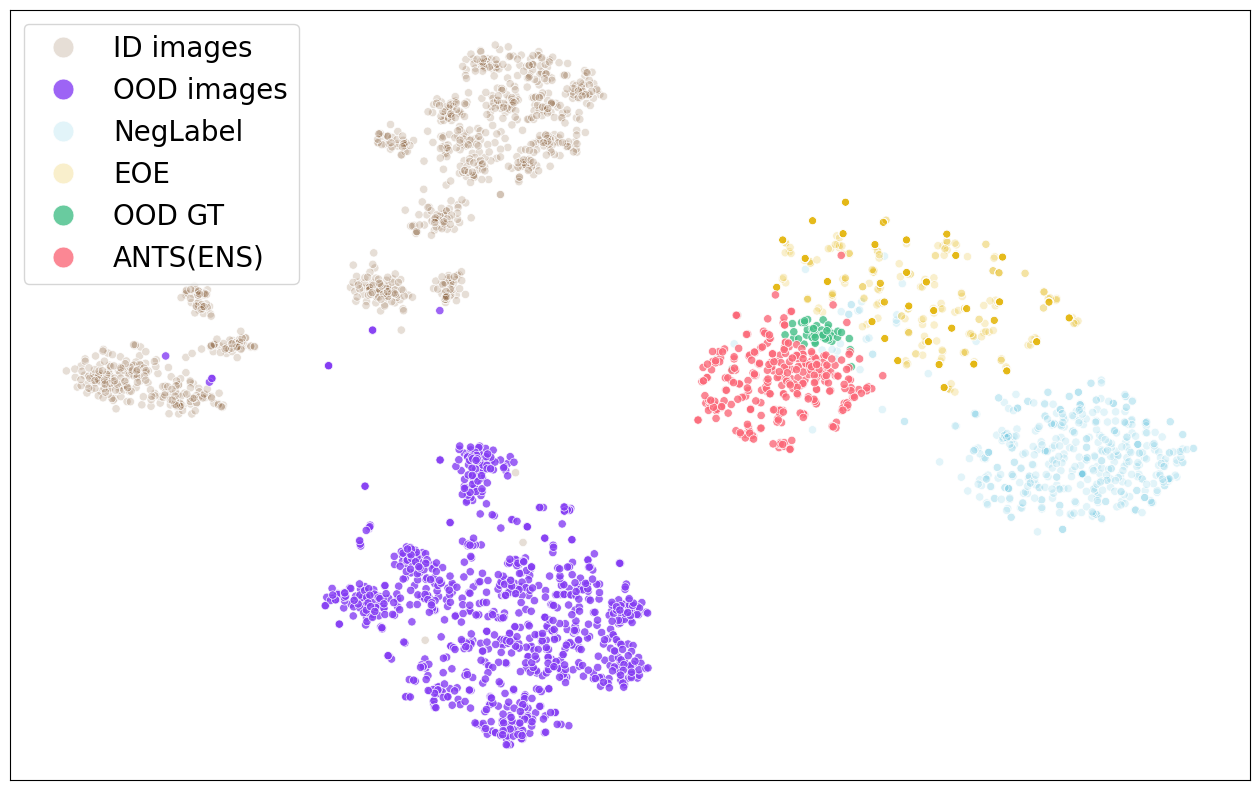}
    \vspace{-0.4cm}
    \caption{T-SNE visualization of the ID and OOD image features, the text features of NegLabel \cite{jiang2024negative}, EOE \cite{cao2024envisioning}, OOD ground-truth, and the expressive negative sentences (ENS) of ANTS. We select ImageNet and SUN as the ID and OOD datasets, respectively. NegLabel and EOE lack a good understanding of OOD images, resulting in a greater distance between the OOD images and the text features. In contrast, our ANTS utilizes the MLLMs to understand OOD images during ENS generation, reducing the distance between ENS and OOD images and improving OOD detection performance.}
    \label{fig:ants_tsne}
    \vspace{-0.4cm}
\end{figure}

\vspace{-0.3cm}
\section{Introduction}
Deep neural networks (DNNs) have achieved remarkable performance in classifying test samples that fall into the training distribution \cite{he2016deep, dosovitskiy2020image}. However, it is well-known that DNNs tend to misclassify test samples from unknown classes, which are often called out-of-distribution (OOD) data \cite{hendrycks2016baseline}. Unfortunately, OOD data are inevitably encountered in open environments. Therefore, how to effectively identify OOD data is crucial for the reliable deployment of DNN models in open-world scenarios.

Traditional OOD detection methods in the image domain primarily rely on visual modality information \cite{ wang2021energy, huang2021importance, sun2021react, wang2021can}. 
For example, MSP~\cite{hendrycks2016baseline} utilizes the maximum softmax probability of a pre-trained vision model to detect OOD images. 
Recently, multimodal knowledge has attracted increasing attention in OOD detection \cite{ming2022delving,miyai2024locoop,li2024learning,nie2024out,bai2024id, jiang2024negative, zhang2024lapt, yu2024self, zhang2024adaneg}. In particular, NegLabel~\cite{jiang2024negative} introduces negative labels (NLs) by mining words that are semantically distant from in-distribution (ID) labels, and identifies OOD images by comparing their similarities to NLs and ID labels. 
Similar approaches generate NLs by prompting LLMs \cite{cao2024envisioning} or modifying superclass names \cite{chen2024conjugated}.
Although these methods have achieved promising performance, they suffer from three key limitations. First, due to the lack of understanding of OOD images, the NLs are positioned far from the OOD image, as illustrated in Fig. \ref{fig:ants_tsne}.
Second, these methods struggle with the challenging near-OOD setting, where OOD samples are semantically close to ID labels. NegLabel focuses on generating NLs that are semantically distant from ID classes, inherently overlooking such cases. While EOE \cite{cao2024envisioning} introduces visually similar labels for all ID classes to address this problem, it neglects the fact that OOD samples are typically similar to only a subset of ID classes, resulting in many false negative labels (see Fig. \ref{fig:vis_false_negative_labels}). 
Third, these methods rely on the strong assumption that the target task setting (\eg, near OOD or far OOD) is known in advance, allowing for the tailored design of NL generation rules. However, this assumption limits their applicability in complex, unknown, and dynamically changing open environments.

\begin{figure}[t]
    \centering
    \includegraphics[width=\linewidth]{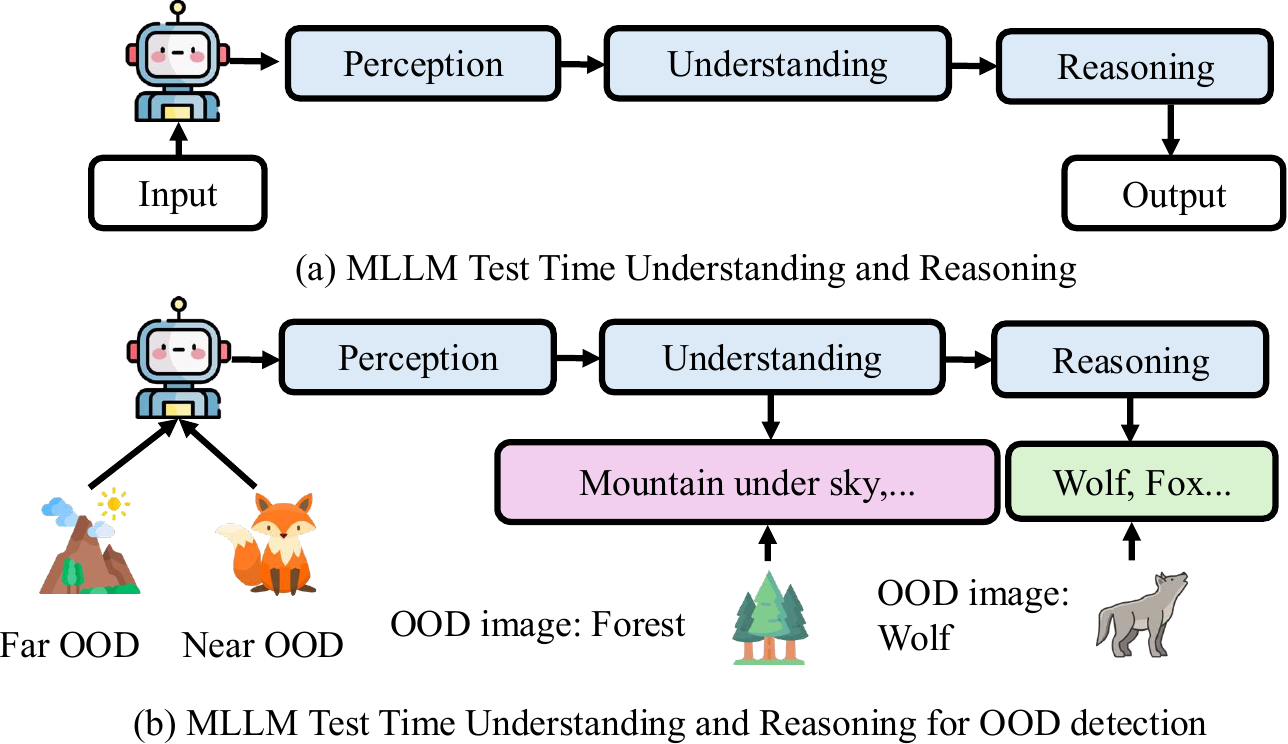}
    \vspace{-0.4cm}
    \caption{(a) Current MLLM improve their reasoning abilities by test time understanding and reasoning through chain-of-thought (CoT) prompting. (b) In our work, we leverage the test time understanding and reasoning capabilities of MLLM during inference to help visual-language models perform better on OOD detection.}
    \label{fig:cot_ood}
    \vspace{-0.4cm}
\end{figure}

\begin{figure*}[t]
    \centering
    \includegraphics[width=0.98\textwidth]{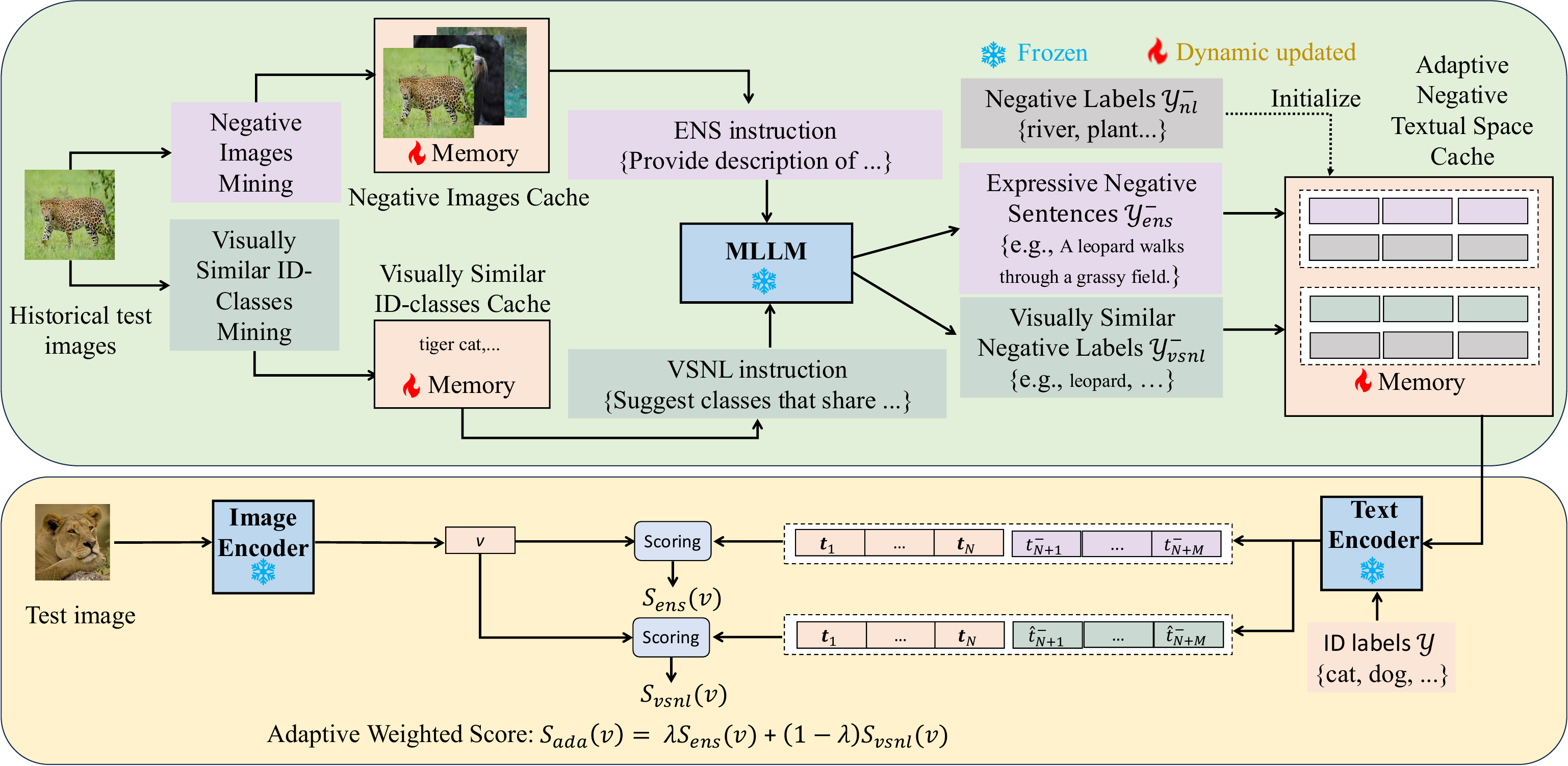}  
        \vspace{-0.2cm}
    \caption{The overall framework of our ANTS. 
    ANTS framework consists of in three stages: (1) caching negative images and visually similar ID classes mined from historical test images; (2) shaping two negative textual spaces by prompting an MLLM with the cached data to generate expressive negative sentences and visually similar labels; and (3) performing online evaluation of the test image using an adaptively weighted combination of these textual spaces.
    }
    \label{fig:framework}
    \vspace{-0.3cm}
\end{figure*}


To address these challenges, we propose to shape an Adaptive Negative Textual Space (ANTS) by harnessing the understanding and reasoning capabilities of multimodal large language models (MLLMs) \cite{liu2023visual, li2023blip, bai2025qwen3}, as shown in Fig. \ref{fig:cot_ood}. Specifically, we introduce expressive negative sentences (ENS), which effectively capture fine-grained details of OOD images. These negative sentences are generated by prompting MLLMs to describe online-mined negative images, leveraging their multimodal understanding capabilities and significantly enhancing the traditional far-OOD detection. While ENS shows greater expressive power in identifying far-OOD samples, it faces challenges in handling the near-OOD setting, where OOD samples are semantically close to certain ID classes. To address this, we dynamically identify the subset of ID classes most similar to the negative images and utilize the reasoning capabilities of MLLMs to construct visually similar negative labels (VSNL) tailored for this subset. This targeted approach reduces false negative labels and improves performance in the near-OOD setting (see Fig. \ref{fig:vis_false_negative_labels}).
Finally, to ensure adaptability across diverse task settings in open environments, we introduce an adaptive weighted score function to balance the two distinct negative textual spaces. This dynamic mechanism enables the model to seamlessly handle both near-OOD and far-OOD scenarios without prior knowledge of the task settings.  The overall framework is presented in Fig. \ref{fig:framework}.

We conduct extensive experiments to validate the advantages of our ANTS method. 
On the large-scale ImageNet dataset, our approach significantly reduces FPR95 by 3.1\% and 3.25\% in the far-OOD and challenging near-OOD detection settings, respectively. Moreover, our method operates in a zero-shot and training-free manner, demonstrating strong scalability across different MLLMs. We summarize our contributions as follows:
\begin{itemize}
    \item We identify three limitations of existing NLs-based methods: (1) lack understanding of OOD images; (2) struggle to address the challenging near-OOD setting, where OOD samples are semantically close to ID labels; (3) rely on the strong assumption that the target task setting (\eg, near-OOD or far-OOD) is known in advance.
    \item To overcome these limitations, we propose the ANTS approach by leveraging the understanding and reasoning capabilities of MLLMs. Specifically, we (1) introduce two strategies including Negative Images Mining and Visually Similar ID-Classes Mining to avoid interference from ID noise and generate false negative labels;
    (2) design two types of prompt for MLLMs to generate expressive negative sentences and visually similar negative labels;
    and (3) design an adaptive weighted score to dynamically balance these two text spaces in open environments.
    \item Extensive experiments are conducted to validate the proposed components. Our method demonstrates new state-of-the-art performance on both near-OOD and far-OOD detection tasks. 
    Our method is training-free, zero-shot, and does not require any auxiliary outlier images. 
\end{itemize}

\section{Related Work}

\noindent\textbf{Traditional OOD Detection.}
Traditional OOD detection methods can be categorized into the following groups: (1) classification-based methods~\cite{hendrycks2016baseline, liang2017enhancing, lee2018simple, liu2020energy, sastry2020detecting, zhang2022out, sun2021react, dong2022neural, sun2022dice, lin2021mood, park2023nearest, jiang2023detecting,  wei2022mitigating, huang2021mos, hendrycks2018deep, yu2019unsupervised, papadopoulos2021outlier, ming2022poem} that distinguish ID and OOD samples by designing a score function; 
(2) density-based methods~\cite{zong2018deep, pidhorskyi2018generative, jiang2021revisiting, wang2022vim} that detect OOD samples by evaluating the likelihood or density of test data derived from probabilistic models; 
(3) distance-based methods~\cite{zaeemzadeh2021out, ming2022exploit} that detect OOD samples by measuring their deviation from in-distribution class prototypes. 

\vspace{0.1cm}
\noindent\textbf{OOD Detection with Vision Language Model.}
 VLM-based OOD detection methods can be broadly categorized into two settings: few-shot and zero-shot. Few-shot methods enhance OOD detection by using negative prompts to define boundaries between ID and OOD images~\cite{nie2024out, li2024learning, bai2024id}, or by integrating non-ID or local ID regions for regularization~\cite{miyai2024locoop, lafon2024gallop}. For zero-shot OOD detection, some works~\cite{ming2022delving, zhang2024adaneg, yang2025oodd, kim2025enhanced, zhu2025knowledge} design post-hoc strategies that utilize softmax scores or image feature information during testing. Some methods~\cite{wang2023clipn, zhang2024lapt} leverage auxiliary datasets to strengthen the detection of OOD samples. Other approaches~\cite{esmaeilpour2022zero, park2023nearest, jiang2024negative, cao2024envisioning, chen2024conjugated, dai2023exploring} retrieve negative labels from corpus databases or generating them using LLMs. However, these NLs methods lack understanding of actual OOD images, the semantic gap with OOD images limits their OOD detection capabilities.

 \vspace{-0.2cm}

\section{Preliminary}
\noindent\textbf{OOD Detection Setup.}
Denote by $\mathcal{X}$ the image space and $\mathcal{Y} = \{ y_1, \dots, y_N \}$ the ID label space, with examples $\mathcal{Y} = \{ cat, dog, \dots, bird \}$ and $N$ denoting the total number of classes. Given $ \vx_{in} \in \mathcal{X}$ as the ID random variable and $\vx_{ood} \in \mathcal{X}$ as the OOD random variable, we denote their respective distributions as $\mathcal{P}_{\vx_{in}}$ and $\mathcal{P}_{\vx_{ood}}$.
In closed-set scenarios, a test image $\vx$ is expected to belong to one ID class, \ie, $\vx \in \mathcal{P}_{\vx_{in}}$ and $y \in \mathcal{Y}$, where $y$ is the label of $\vx$.
However, in real-world scenarios, AI systems may encounter samples that do not match any known class, \ie, $\vx \in \mathcal{P}_{\vx_{ood}}$ and $y \notin \mathcal{Y}$, resulting in potential misclassifications and safety concerns \cite{scheirer2012toward}.
To tackle these issues, OOD detection aims to distinguish ID and OOD samples using a scoring function $S$:
\begin{equation} \label{Equ:ood_score_function}
G_\gamma(\vx)= \begin{cases} \text{ID} & S(\vx) \geq \gamma, \\ \text{OOD} & S(\vx) < \gamma, \end{cases}
\end{equation}
where $G_{\gamma}$ is the OOD detector with threshold $\gamma$.

\vspace{0.1cm}
\noindent\textbf{OOD Detection with NLs.}  
Enhancing OOD detection with textual knowledge has recently garnered increasing attention \cite{ming2022delving,wang2023clipn,zhang2024adaneg}, while a representative type of approach introduces NLs \cite{jiang2024negative,cao2024envisioning}.
Specifically, in addition to the ID labels $\mathcal{Y}$, these methods introduce a disjoint set of NLs $\mathcal{Y}^-$ and classify a test sample as OOD if it exhibits high similarity to NLs and low similarity to ID labels. In this process, the quality of NLs is crucial.  
The pioneering method, NegLabel \cite{jiang2024negative}, selects words with large cosine distance to ID labels in a large corpus dataset $\mathcal{Y}^{c} =  \left\{ \widetilde{y}_1, \widetilde{y}_2, \ldots, \widetilde{y}_K \right\}$ as NLs:  
\begin{equation} \label{eq:neglabel_texts}
    \mathcal{Y}^-_{nl} = \mathcal{G}_{dis}(\mathcal{Y}, \mathcal{Y}^{c}, f_{clip}, M),
\end{equation}  
where the CLIP-like model $f_{clip}$ defines the label similarity space. $K$ and $M$ represent the numbers of candidate labels in $\mathcal{Y}^{c}$ and the selected NLs in $\mathcal{Y}^-_{nl}$, where $M \leq K$.  
Another representative work, EOE \cite{cao2024envisioning}, uses prompts to guide an LLM to generate NLs:  
\begin{equation}
    \mathcal{Y}^-_{eoe} = \mathcal{G}_{llm}(\mathcal{Y}, f_{llm}, \rho_{neg}, M),
\end{equation}  
where $\rho_{neg}$ is a carefully designed textual prompt for the LLM $f_{llm}$. Given the generated NLs (\eg, $\mathcal{Y}^-_{nl}$ \cite{jiang2024negative}), the score function for OOD detection can be formulated as:
\begin{equation}
    S_{nl}(\vv) =  \frac{\sum_{y \in \mathcal{Y}} {e^{\cos(\vv, \vt) / \tau}}}{\sum_{y \in \mathcal{Y}} {e^{\cos(\vv, \vt)/ \tau}} + \sum_{y^- \in \mathcal{Y}^-_{nl}} {e^{\cos(\vv, \vt^-)/ \tau}}},
\end{equation}
where $\tau > 0$ is the temperature scaling parameter.
$\vv \in \mathcal{R}^D$ represents the test image feature, while $\vt \in \mathcal{R}^D$ and $\vt^- \in \mathcal{R}^D$ denote the text features of ID labels $y \in \mathcal{Y}$ and NLs $y^- \in \mathcal{Y}^-_{nl}$, respectively, where $D$ is the feature dimension.

\section{Methodology}

\subsection{Motivation}
Although NegLabel~\cite{jiang2024negative} and EOE~\cite{cao2024envisioning} have advanced OOD detection using NLs, they face three key limitations: (1) lacking of understanding of OOD images, as shown in Fig.\ref{fig:ants_tsne}; (2) poor performance in near-OOD settings due to false negatives by neglecting visually similar classes; and (3) reliance on prior task knowledge, limiting adaptability in open environments. This motivates us to raise the following question: 

\textit{Can we leverage the test time understanding and reasoning capabilities of MLLMs to shape a more accurate and comprehensive negative textual space?}

In this work, we attempt to answer this question by designing different prompts for MLLMs to leverage their test-time understanding and reasoning capabilities for OOD detection, as shown in Fig.\ref{fig:cot_ood}. The overall pipeline of our method is illustrated in Fig. \ref{fig:framework}.

\subsection{Expressive Negative Sentences}
\begin{figure*}[t] 
   \begin{subfigure}[t]{0.32\textwidth}
\includegraphics[width=\linewidth]{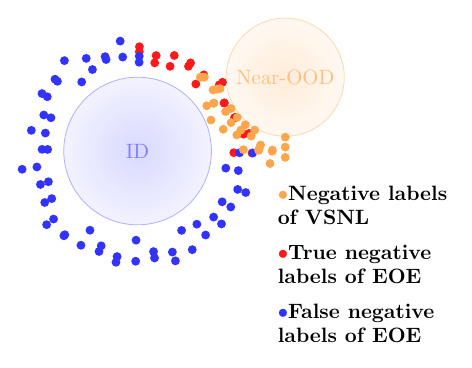}
        \caption{False Negative NLs}
        \label{fig:vis_false_negative_labels}
\end{subfigure}
\hfill 
    \begin{subfigure}[t]{0.32\textwidth}
\includegraphics[width=\linewidth]{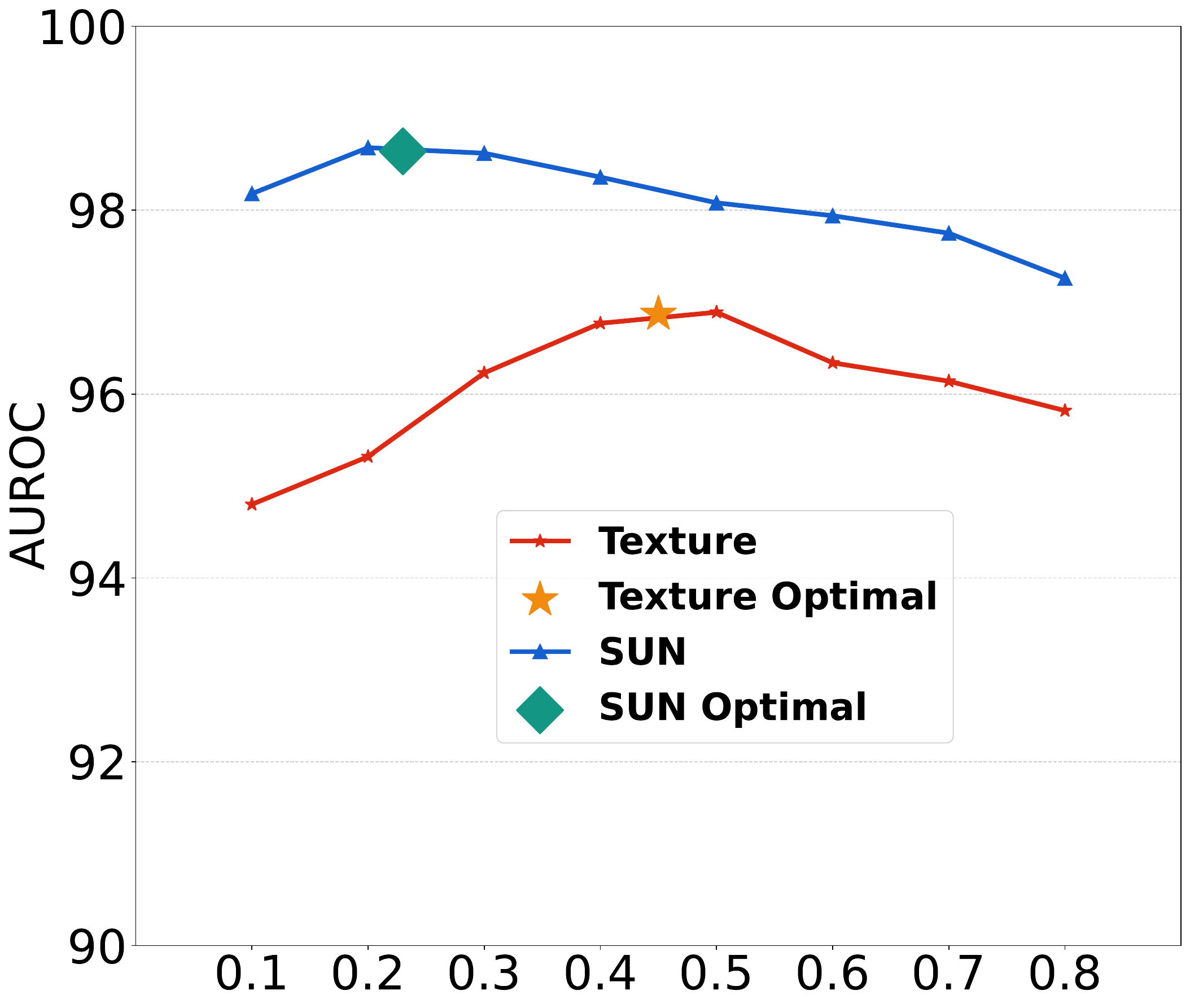}
        \caption{Different optimal thresholds $\gamma$
        }
        \label{fig:opt_threshold}
    \end{subfigure}
    \hfill 
    \begin{subfigure}[t]{0.32\textwidth} 
\includegraphics[width=\linewidth]{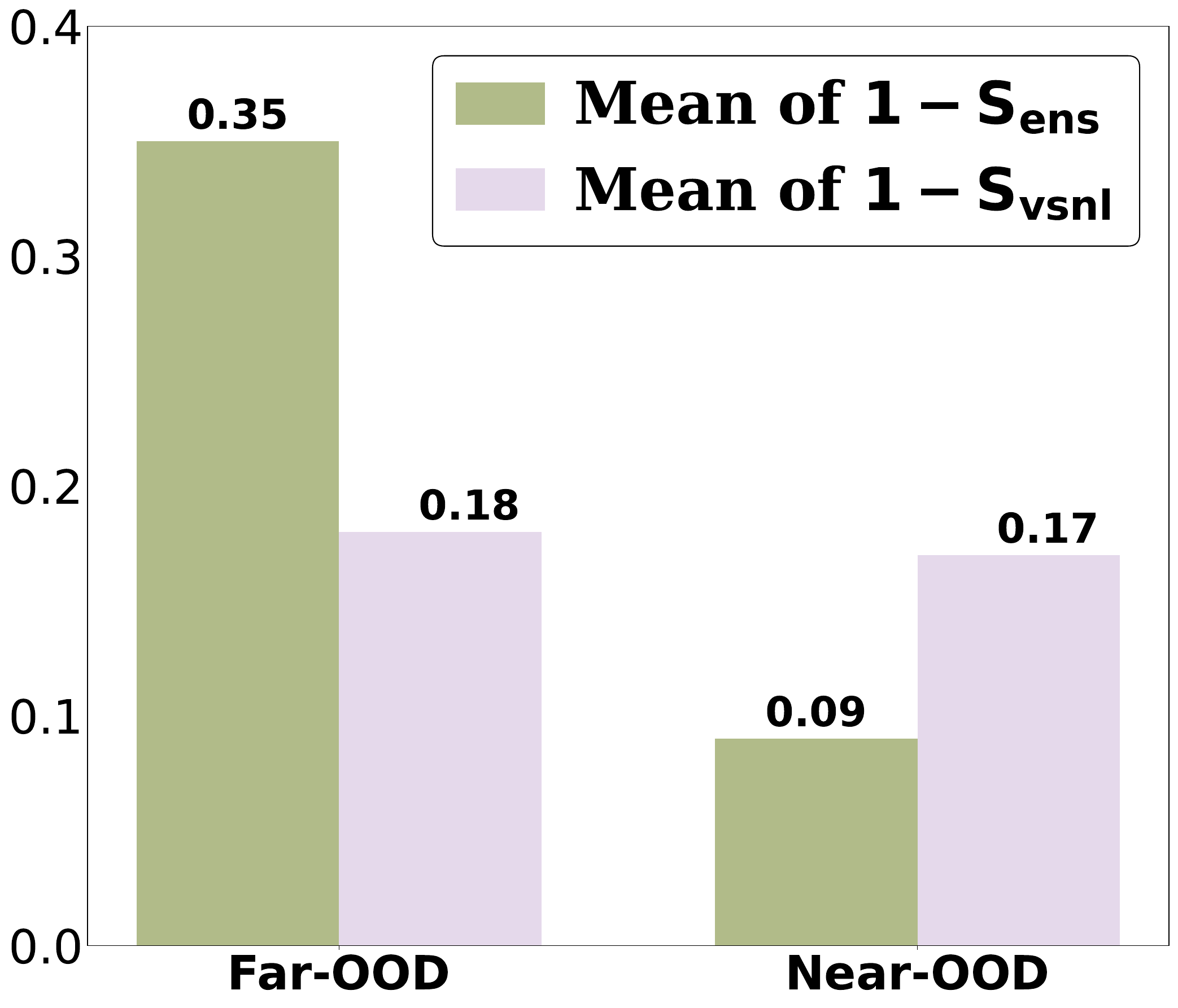}
        \caption{$1-S_{ens}(\vv)$ Vs. $1-S_{vsnl}(\vv)$
        }
        \label{fig:score_x_neg_u}
    \end{subfigure}  
    \caption{
(a) Our VSNL generates visually similar labels only for the ID class subset, whose images are most similar to the near OOD samples, largely reducing false negative labels. 
(b) Different OOD datasets prefer different thresholds, and our proposed method can cache the historical test images and adaptively mine negative images, implicitly setting an dataset adaptive threshold. 
(c) $S_{ens}$ and $S_{vsnl}$ perform differently on far and near OOD, providing clues for designing an adaptive weight $\lambda$ in Eq. \ref{eq:ada_score}.
    }
    \label{fig:trend}
    \vspace{-0.3cm}
\end{figure*}

\textbf{Negative Images Mining.} We leverage the image understanding capabilities of MLLMs to generate expressive negative sentences by describing negative images, which are historical test images likely to be OOD samples.  
We identify these negative images using the OOD detector of NegLabel, where historical test images with $\mathcal{S}_{nl}(\vx) < \gamma$ are selected as negative images: 
\begin{equation} \label{eq:neglabel_negmine}
    \mathcal{X}_{neg} = \{ \vx \mid S_{nl}(\vx) < \gamma, \ \vx \in \mathcal{X}^{\rm his}_{test} \},
\end{equation}  
where $\mathcal{X}^{\rm his}_{test}$ denotes the historical test data. 
We find that manually defining a fixed $\gamma$ is challenging for handling different testing scenarios, as the optimal threshold varies between different OOD datasets, as analyzed in Fig. \ref{fig:opt_threshold}. To address this issue, we develop an adaptive threshold determination strategy based on the characteristics of the historical test data. Specifically, we filter out historical test samples with high $\mathcal{S}_{nl}$ scores using Eq. \ref{eq:neglabel_negmine}, as these samples are highly likely to be ID samples.
For the remaining negative images $\widehat{\mathcal{X}}_{neg}$, which fall into a mixed set of ID and OOD samples, we select a proportion $\eta$ of images with the lowest $S_{nl}$ scores, and the adaptive threshold $\gamma^{*}$ can be formulated as:
\begin{align} \label{eq:ens_percentile_split}
    \mathcal{X}_{neg} &= \text{Top}(\widehat{\mathcal{X}}_{neg}, \mathcal{O}_{nl}, \eta), 
    \gamma^{*} &= \max_{\mathbf{x} \in \mathcal{X}_{neg}} S_{nl}(\mathbf{x}),
\end{align}
where $\mathcal{O}_{nl} = \{-S_{nl}(\vx) \mid \vx \in \widehat{\mathcal{X}}_{neg} \}$, and $\eta \in (0,1)$ determines the selection ratio. Here, function $\text{Top}(A, B, \eta)$ selects a proportion of $\eta$ indices with the highest values in set $B$, and then retrieves the corresponding images from set $A$ based on these indices. Since this approach relies on the distribution of $\mathcal{O}_{nl}$, it is equivalent to using an adaptive, data-dependent threshold $\gamma$, as illustrated in Fig.~\ref{fig:opt_threshold}.

\begin{figure}[t]
\centering
\includegraphics[width=\linewidth]{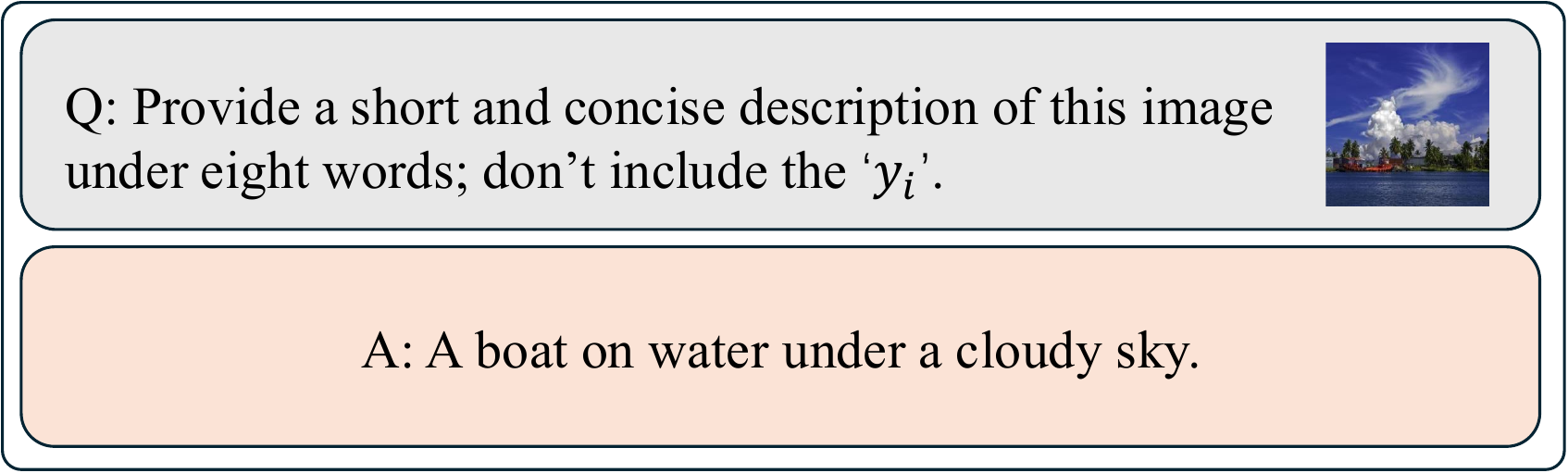}  
\caption{Expressive Negative Sentences, where $y_{i}$ represents the predicted ID label of the negative image.}
\label{fig:ins_ens}
\end{figure}

\begin{figure}[t]
\includegraphics[width=\linewidth]{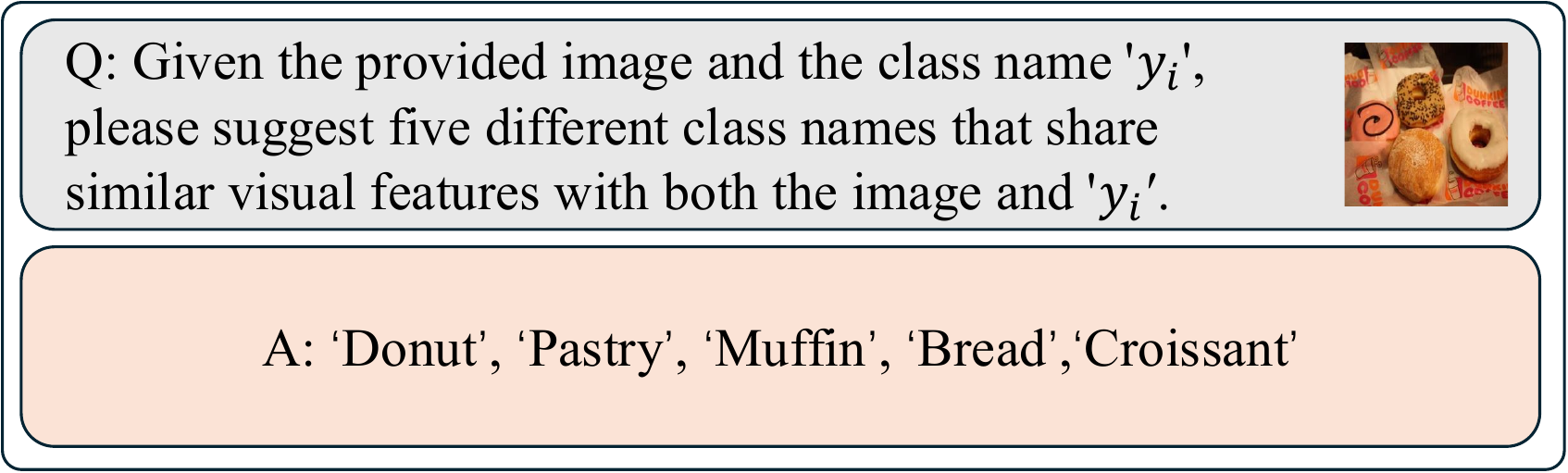}  
\caption{Visually Similar Negative Labels, where $y_{i}$ represents the predicted ID label of the negative image.}
\vspace{-0.2cm}
\label{fig:ins_nvsl}
\end{figure}

\vspace{0.1cm}
\noindent\textbf{ENS Generation and Score.}
With the mined negative images $\mathcal{X}_{neg}$, we introduce the expressive negative sentences as follows:
\begin{equation} \label{eq:ens}
    \mathcal{Y}^-_{ens} = \mathcal{G}_{ens}(\mathcal{Y}, \mathcal{X}_{neg}, f_{mllm}, M),
\end{equation}  
where $\mathcal{G}_{ens}$ is the negative sentence generation process detailed in Fig. \ref{fig:ins_ens}. If $|\mathcal{X}_{neg}| \geq M$, we randomly select $M$ sentences. Otherwise, we repeat the prompting process to generate $M$ negative sentences.
With the expressive negative sentences, we introduce the following negative score:
\begin{equation} \label{eq:ens_score}
    S_{ens}(\vv) =  \frac{\sum_{y \in \mathcal{Y}} {e^{\cos(\vv, \vt) / \tau}}}{\sum_{y \in \mathcal{Y}} {e^{\cos(\vv, \vt)/ \tau}} + \sum_{y^- \in \mathcal{Y}^-_{ens}} {e^{\cos(\vv, \vt^-)/ \tau}}}.
\end{equation}

\subsection{Visually Similar Negative Labels}
The expressive negative sentences introduced above can enhance the detection of far-OOD samples by describing negative images in detail. However, they struggle to distinguish ID samples from visually similar near-OOD data, as both conform to the sentence descriptions. To address this limitation, we prompt the MLLM to generate visually similar labels for the ID labels:  
\begin{equation} \label{eq:vsim}
    \mathcal{Y}^-_{vsl} = \mathcal{G}_{vsnl}(\mathcal{Y}, \mathcal{X}^{\rm his}_{test}, f_{mllm}, M),
\end{equation}  
where $\mathcal{G}_{vsnl}$ represents the visually similar label generation process, as illustrated in Fig.~\ref{fig:ins_nvsl}.  

\vspace{0.1cm}
\noindent\textbf{Visually Similar ID-Classes Mining.}
While these visually similar labels cover the near-OOD regions, they typically include false NLs. Specifically, OOD data may only be similar to a subset of ID classes, while being distant from others. 
These visually similar NLs derived from OOD-unrelated ID classes are also far from OOD samples, thereby introducing false NLs and disturbing OOD detection, as intuitively shown in Fig. \ref{fig:vis_false_negative_labels}.  To address this issue, we first identify the subset of ID labels most similar to the OOD samples:  
\begin{equation}
\label{eq:subset_id_similar}
\begin{aligned} 
    F(y_{i}) &= \frac{|\{\vx \in \mathcal{X}^{\rm his}_{test} \mid H(\vx) = y_{i}\}|}{|\mathcal{X}^{\rm his}_{test}|}, \quad \forall y_{i} \in \mathcal{Y}, \\
    \quad \mathcal{Y}' &= \text{Top}\left(\mathcal{Y}, F(\mathcal{Y}), \delta\right),
\end{aligned}
\end{equation}
where $H(\vx)$ is the CLIP-based ID classifier with ID text features as weight, $|\cdot|$ measures the set size,
$F(y_i)$ represents the proportion of historical test images in $\mathcal{X}^{\rm his}_{test}$ being classified as $y_i \in \mathcal{Y}$, $F(\mathcal{Y})$ is the collection of $F(y_i)$, and 
$\delta \in (0, 1)$ serves as the selection ratio.

\vspace{0.1cm}
\noindent\textbf{VSNL Generation and Score.} 
After getting these filtered ID labels that share high similarity with negative images, we introduce the following visually similar negative labels:
\begin{equation} \label{eq:vsnl}
    \mathcal{Y}^-_{vsnl} = \mathcal{G}_{vsnl}(\mathcal{Y}', \mathcal{X}^{\rm his}_{test}, f_{mllm}, M).
\end{equation}  
These visually similar negative labels adaptively capture the characteristics of the target OOD distribution, reducing false negative labels and resulting in the following score function:
\begin{equation} \label{eq:vsnl_score}
    S_{vsnl}(\vv) =  \frac{\sum_{y \in \mathcal{Y}} {e^{\cos(\vv, \vt) / \tau}}}{\sum_{y \in \mathcal{Y}} {e^{\cos(\vv, \vt)/ \tau}} + \sum_{y^- \in \mathcal{Y}^-_{vsnl}} {e^{\cos(\vv, \widehat{\vt}^-)/ \tau}}},
\end{equation}
where $\widehat{\vt}^-$ is the text feature of $y^- \in \mathcal{Y}^-_{vsnl}$.

\subsection{Adaptive Weighted Score}
Existing OOD detection methods rely on the assumption that the testing scenario (near-OOD or far-OOD) is human defined beforehand, but in real-world applications, this assumption often fails due to the dynamic nature of open environments.
To address this, we propose an adaptive weighting strategy to balance these two  scoring functions with an adaptive weight $\lambda \in [0,1]$:  
\begin{equation} \label{eq:ada_score}
    S_{ada}(\vv) = \lambda S_{ens}(\vv) + (1-\lambda) S_{vsnl}(\vv).
\end{equation}  
The weight $\lambda$ adjusts dynamically based on the environment, approaching $1$ in far-OOD scenarios to prioritize $S_{ens}(\vv)$, and $0$ in near-OOD scenarios to emphasize $S_{vsnl}(\vv)$.

We design the adaptive weight $\lambda$ by leveraging the performance differences of $S_{ens}(\vv)$ and $S_{vsnl}(\vv)$ on near and far OOD data. Specifically, ENS effectively characterizes far OOD samples, but its coarse-grained descriptions struggle to distinguish near OOD from ID samples, resulting in lower scores for far OOD samples and higher scores for near OOD samples. Conversely, VSNL better captures near OOD samples but, due to its ID-similarity, produces false negatives for far OOD samples, leading to higher scores for far OOD samples and lower scores for near OOD samples, as illustrated in Fig. \ref{fig:score_x_neg_u}. 
Based on this observation, we define $\lambda$ as:
\begin{align} \label{eq:adaptive_weight}
    \lambda &= F\left(
    \frac{1}{|\mathcal{X}_{neg}|} \sum_{\vv \in \mathcal{X}_{neg}}S_{ens}(\vv), \,
    \frac{1}{|\mathcal{X}_{neg}|} \sum_{\vv \in \mathcal{X}_{neg}}S_{vsnl}(\vv)
    \right), 
\end{align}
where $F(a, b) = \frac{1 - a}{(1 - a) + (1 - b)} \in (0,1)$. One can see that when $\frac{1}{|\mathcal{X}_{neg}|} \sum_{\vv \in \mathcal{X}_{neg}}S_{ens}(\vv) > \frac{1}{|\mathcal{X}_{neg}|} \sum_{\vv \in \mathcal{X}_{neg}}S_{vsnl}(\vv)$, $\lambda$ approaches 0; otherwise, $\lambda$ approaches 1. 
The algorithm is summarized in Alg.~\ref{alg:ants}.

\begin{table*}[t] 
\small
\setlength{\tabcolsep}{3pt}
\centering
\caption{OOD detection results by using ImageNet-1k as ID dataset. ViTB/16 is used as the encoder. The results of traditional methods are available in the \textbf{supplementary materials}.} 
\vspace{-0.2cm}
\begin{tabular}{lcccccccc|cc}
\toprule
\multicolumn{11}{c}{OOD datasets}  \\
\multicolumn{1}{c}{{Methods}} & \multicolumn{2}{c}{INaturalist} & \multicolumn{2}{c}{SUN} & \multicolumn{2}{c}{Places} & \multicolumn{2}{c}{Textures} & \multicolumn{2}{c}{Average} \\ \cline{2-3} \cline{4-5} \cline{6-7} \cline{8-9} \cline{10-11}
 & \fontsize{8}{12}\selectfont AUROC$\uparrow$ & \fontsize{8}{12}\selectfont FPR95$\downarrow$& \fontsize{8}{12}\selectfont AUROC$\uparrow$ & \fontsize{8}{12}\selectfont FPR95$\downarrow$& \fontsize{8}{12}\selectfont AUROC$\uparrow$ & \fontsize{8}{12}\selectfont FPR95$\downarrow$& \fontsize{8}{12}\selectfont AUROC$\uparrow$ & \fontsize{8}{12}\selectfont FPR95$\downarrow$ & \fontsize{8}{12}\selectfont AUROC$\uparrow$ & \fontsize{8}{12}\selectfont FPR95$\downarrow$  \\
 \midrule
  \multicolumn{11}{c}{\textbf{Training-required (or with Fine-tuning)}} \\ 
MSP~\cite{hendrycks2016baseline} & 87.44 & 58.36 & 79.73 & 73.72 & 79.67 & 74.41 & 79.69 & 71.93 & 81.63 & 69.61   \\
ZOC~\cite{esmaeilpour2022zero} & 86.09 & 87.30 & 81.20 & 81.51 & 83.39 & 73.06 & 76.46 & 98.90 & 81.79 & 85.19 \\
CLIPN~\cite{wang2023clipn} & 95.27 & 23.94 & 93.93 & 26.17 & 92.28 & 33.45 & 90.93 & 40.83 & 93.10 & 31.10 \\
LSN~\cite{nie2024out} & 95.83 & 21.56 & 94.35 & 26.32 & 91.25 & 34.48 & 90.42 & 38.54 & 92.26 & 30.22 \\
LoCoOp~\cite{miyai2024locoop} & 93.93 & 29.45 & 90.32 & 41.13 & 90.54 & 44.15 & 93.24 & 33.06 & 92.01 & 36.95 \\
ID-Like~\cite{bai2024id} & 98.19 & 8.98 & 91.64 & 42.03 & 90.57 & 44.00 & 94.32 & 25.27 & 93.68 & 30.07 \\
NegPrompt~\cite{li2024learning} & 90.49 & 37.79 & 92.25 & 32.11 & 91.16 & 35.52 & 88.38 & 43.93 & 90.57 & 37.34 \\
SCT~\cite{yu2024self} & 95.86 & 13.94 & 95.33 & 20.55 & 92.24 & 29.86 & 89.06 & 41.51 & 93.27 & 26.47 \\
LAPT~\cite{zhang2024lapt} & 99.63 & 1.16 & 96.01 & 19.12 & 92.01 & 33.01 & 91.06 & 40.32 & 94.68 & 23.40 \\
CMA~\cite{kim2025enhanced} & 99.62 & 1.65 & 96.36 & 16.84 & 93.11 & 27.65 & 91.64 & 33.58 & 95.13 & 19.93 \\
SynOOD~\cite{li2025synthesizing} & 99.57 & 1.57 & 95.82 & 20.46 & 97.37 & 12.12 & 95.29 & 22.94 & 97.01 & 14.27 \\
\midrule
\multicolumn{11}{c}{\textbf{Zero Shot (No Training Required)}} \\
MCM~\cite{ming2022delving} & 94.59 & 32.20 & 92.25 & 38.80 & 90.31 & 46.20 & 86.12 & 58.50 & 90.82 & 43.93 \\
CoVer~\cite{zhang2024if} & 95.98 & 22.55 & 93.42 & 32.85 & 90.27 & 40.71 & 90.14 & 43.39 & 92.45 & 34.88 \\
EOE~\cite{cao2024envisioning} & 97.52 & 12.29 & 95.73 & 20.40 &  92.95 & 30.16 & 85.64 & 57.63 & 92.96 & 30.09 \\
NegLabel~\cite{jiang2024negative} & 99.49 & 1.91 & 95.49 & 20.53 & 91.64 & 35.59 & 90.22 & 43.56 & 94.21 & 25.40 \\
OODD~\cite{yang2025oodd} & 99.36 & 2.22 & 95.01 & 21.49 & 87.10 & 44.76 & 93.27 & 30.69 & 93.69 & 24.79 \\
AdaNeg~\cite{zhang2024adaneg} & 99.71 & 0.59 & 97.44 & 9.50 & 94.55 & 34.34 & 94.93 & 31.27 & 96.66 & 18.92 \\
CSP~\cite{chen2024conjugated} & 99.60 & 1.54 & 96.66 & 13.66 & 92.90 & 29.32 & 93.86 & 25.52 & 95.76 & 17.51 \\
\rowcolor{lightpink}
\textbf{ANTS} & \textbf{99.75} & \textbf{0.54} & \textbf{98.77} & \textbf{5.43} & 96.10 & 20.21 & \textbf{96.38} & \textbf{18.52} & \textbf{97.75} & \textbf{11.20} \\
\bottomrule
\end{tabular}
\label{tab:traditional_four_ood_datasets}
\end{table*}

\begin{algorithm}[t]
\caption{Adaptive Negative Textual Space Shaping}
\label{alg:ants}
\begin{algorithmic}[1]
\REQUIRE ID label space $\mathcal{Y}$, stream of testing batches $\{\mathcal{X}_t\}_{t=1}^T$.
\STATE Initialize ENS space $\mathcal{Y}^-_{ens}$ and VSNL space $\mathcal{Y}^-_{vsnl}$ with $\mathcal{Y}^{-}_{nl}$ of Eq. \ref{eq:neglabel_texts}.
\FOR{each incoming batch $\mathcal{X}_t$}
    \STATE Filter historical test samples with $\mathcal{S}_{nl}$ using Eq. \ref{eq:neglabel_negmine};  \hfill
    \STATE Collect negative images $\mathcal{X}_{neg}$ with implicit and adaptive threshold using Eq. \ref{eq:ens_percentile_split};  \hfill
    \STATE Generate expressive negative sentences $\mathcal{Y}^-_{ens}$ with the MLLM using Eq. \ref{eq:ens}; \hfill \textcolor{gray}{// ENS Generation} 
    \STATE Identify ID label subset similar to historical test images using Eq. \ref{eq:subset_id_similar};
    \STATE Generate visually similar negative labels $\mathcal{Y}^-_{vsnl}$ with MLLM using Eq. \ref{eq:vsnl}; \hfill \textcolor{gray}{// VSNL Generation}
    \STATE Compute ENS scores $S_{ens}$ and VSNL scores $S_{vsnl}$ using Eq. \ref{eq:ens_score} and Eq. \ref{eq:vsnl_score}, respectively;
    \STATE Calculate the dynamic weighting $\lambda$ using Eq. \ref{eq:adaptive_weight};
    \STATE Get the final score $S_{ada}$ by weighting the two scores using Eq. \ref{eq:ada_score}. \hfill
    \textcolor{gray}{//Adaptive Weighted Score}
\ENDFOR
\STATE \textbf{Output} Collected score $S_{ada}$.
\end{algorithmic}
\vspace{-0.1cm}
\end{algorithm}

\section{Experiments}
\vspace{-0.2cm}
\subsection{Experiment Setup} \label{subsec:exp_setup}
\vspace{-0.1cm}
\noindent\textbf{Datasets and benchmarks.}
Following \cite{huang2021importance}, we select ImageNet-1K~\cite{deng2009imagenet} as the ID dataset and use iNaturalist~\cite{van2018inaturalist}, SUN~\cite{xiao2010sun}, Places~\cite{zhou2017places}, and Textures~\cite{cimpoi2014describing} as the OOD test datasets. We also validate our method on the OpenOOD benchmark, which contains SSB-hard~\cite{vaze2021open} and NINCO~\cite{bitterwolf2023or} as near-OOD datasets, and
iNaturalist~\cite{van2018inaturalist}, Texture~\cite{cimpoi2014describing}, and OpenImage-O~\cite{wang2022vim} as far-OOD datasets.

\noindent\textbf{Implementation Details.}
We use the visual encoder of ViT-B/16 pretrained by CLIP~\cite{radford2021learning}. We adopt the LLaVA-1.5-7B model as the default MLLM for our research,
Following NegLabel~\cite{jiang2024negative}, we adopt the text prompt of ‘The nice <label>.’, set temperature $\tau$ =0.01, and define the number $M$ of negative labels as 10000 with group size 100.
Additionally, we set the initial threshold $\gamma = 0.9$ and set $\eta $ = 0.5. Our method follows a test-time adaptation setting as \cite{zhang2024adaneg}. 

\noindent\textbf{Evaluation Metrics.} 
We employ two standard metrics to evaluate OOD detection: the false positive rate (FPR95), which measures the rate of OOD samples when the true positive rate for ID samples is at 95\%, and the area under the receiver operating characteristic curve (AUROC).

\subsection{Main Results}

\begin{table}[t]
\caption{OOD detection results of zero-shot methods on the OpenOOD benchmark. ImageNet-1k is adopted as ID dataset. Detailed results are available
in the \textbf{supplementary materials}.}\label{tab:openood}
\vspace{-0.1cm}
\small
\setlength{\tabcolsep}{3pt}
\centering
\begin{tabular}{l|cc|cc}
\toprule
\multirow{2}{*}{ Methods } & \multicolumn{2}{|c|}{ FPR95 $\downarrow$} & \multicolumn{2}{|c}{$\mathrm{AUROC} \uparrow$} \\
\cline{2-5}
& Near-OOD & Far-OOD & Near-OOD & Far-OOD \\
\midrule
MCM~\cite{ming2022delving} & 79.02 & 68.54 & 60.11 & 84.77 \\
NegLabel~\cite{jiang2024negative} & 68.18 & 27.34 & 76.92 & 93.30 \\
EOE~\cite{cao2024envisioning} & 82.93 & 46.73 & 66.94 & 89.14 \\
AdaNeg~\cite{zhang2022out} & 67.51 & 17.31 & 76.70 & 96.43 \\
SynOOD~\cite{li2025synthesizing} & 71.68 & 17.11 & 77.55 & 96.21 \\
\rowcolor{lightpink} 
\textbf{ANTS} & \textbf{60.98} & \textbf{15.38} & \textbf{82.15} & \textbf{96.50} \\
\bottomrule
\end{tabular}
\vspace{-0.2cm}
\end{table}

\begin{table}[h]
\centering
\small
\caption{ OOD detection performance on other ID datasets.}
\vspace{-0.1cm}
\label{tab:other_ids}
\begin{tabular}{lccccc}
\toprule
ID Dataset & Method & AUROC$\uparrow$ & FPR95$\downarrow$ \\
\midrule
\multirow{2}{*}{CUB-200-2011} 
    & NegLabel \cite{jiang2024negative} & 99.93 & 0.13 \\
    & \textbf{ANTS (Ours)} & \textbf{99.95} & \textbf{0.01} \\
\midrule
\multirow{2}{*}{STANFORD-CARS}
    & NegLabel \cite{jiang2024negative} & 99.99 & 0.01 \\
    & \textbf{ANTS (Ours)} & \textbf{99.99} & \textbf{0.00} \\
\midrule
\multirow{2}{*}{Food-101}
    & NegLabel \cite{jiang2024negative} & 99.90 & 0.40 \\
    & \textbf{ANTS (Ours)} & \textbf{99.92} & \textbf{0.05} \\
\midrule
\multirow{2}{*}{Oxford-IIIT Pet}
    & Neglabel \cite{jiang2024negative} & 99.62 & 1.70 \\
    & \textbf{ANTS (Ours)} & \textbf{99.99} & \textbf{0.02} \\
\bottomrule
\end{tabular}
\vspace{-0.5cm}
\end{table}

\vspace{-0.2cm}


\vspace{0.1cm}
\noindent\textbf{ImageNet-1k Benchmark.}
The results are shown in Tab.~\ref{tab:traditional_four_ood_datasets}. We can see that other CLIP-based training-based methods~\cite{wang2023clipn, nie2024out, miyai2024locoop, bai2024id, li2024learning, yu2024self, zhang2024lapt} learn a negative branch or prompts by synthesizing negative samples, but they often fail to reflect the true OOD space. Other NL-based methods~\cite{jiang2024negative, cao2024envisioning, chen2024conjugated} directly retrieve negative labels from corpus datasets or generate them using LLMs, but they lack supervision from auxiliary OOD images. ANTS consistently achieves remarkable improvements, which demonstrates the advantages of utilizing the understanding capabilities of MLLMs to shape a more accurate NL space. Compared with other test-time adaptation methods ~\cite{zhang2024adaneg, yang2025oodd}, which store test images in memory to calculate the image proxy score and then combine the scores from both modalities, ANTS uses a text-only score to eliminate the modality gap when calculating the OOD score with ID classes, leading to better OOD detection results.

\noindent\textbf{OpenOOD Benchmark.}
The results are shown in Tab. \ref{tab:openood}. Though the NL-based methods~\cite{jiang2024negative, cao2024envisioning} can handle small-scale near-OOD scenarios (\eg, using ImageNet-10 and ImageNet-20 as ID and OOD data, respectively), these methods that selects semantically distant negative labels struggle to handle large-scale near-OOD scenarios such as using ImageNet-1k as ID. However, EOE~\cite{cao2024envisioning}, while using LLMs to generate visually similar labels, suffers from a growing number of false negatives with increasing ID classes. ANTS first identifies a subset of ID classes similar to OOD images, as shown in Fig.~\ref{fig:vis_false_negative_labels}, then it leverages the reasoning capabilities of MLLMs to generate visually similar labels. As a result, ANTS significantly outperforms its closest competitors~\cite{jiang2024negative,cao2024envisioning, zhang2024adaneg} in both near-OOD and far-OOD scenarios, validating its scalability.

\noindent\textbf{Results of other ID datasets.} 
As shown in Tab.~\ref{tab:other_ids}, our ANTS consistently surpasses existing methods in zero-shot OOD detection method NegLabel\cite{jiang2024negative} across all in-distribution (ID) datasets. We also validate the robustness of ANTS to \textbf{Domain Shift} and \textbf{Adversarial Examples}, the detailed results are available in the supplementary materials.

\begin{table}[t]
\centering
\caption{Ablation experiments. `NIM' indicates the Negative Image Mining strategy in Eq.~\ref{eq:ens_percentile_split}, and `SIM' means the Visually Similar ID-Classes Mining strategy in Eq.~\ref{eq:subset_id_similar}.  }
    \scriptsize
    \begin{tabular}{@{}c|cc|cc|c|c|c@{}}
    \toprule
    & \multicolumn{5}{c|}{Components} & \multicolumn{2}{c}{FPR95 $\downarrow$}  \\
   & NIM & $\mathcal{Y}^-_{ens}$ & SIM & $\mathcal{Y}^-_{vsnl}$ & $S_{ada}(\vv)$  & NearOOD  & FarOOD  \\
    \midrule
     \multicolumn{6}{c|}{NegLabel\cite{jiang2024negative}} & 68.18 & 27.34  \\  
        \midrule
    A & \textcolor{gray}{\XSolidBrush} &  \Checkmark & \textcolor{gray}{\XSolidBrush} & \textcolor{gray}{\XSolidBrush} & \textcolor{gray}{\XSolidBrush} & \textcolor{gray}{74.48} & 43.87 \\
    B & \Checkmark &  \Checkmark & \textcolor{gray}{\XSolidBrush} & \textcolor{gray}{\XSolidBrush} & \textcolor{gray}{\XSolidBrush} & \textcolor{gray}{73.70} & 19.22 \\
    \midrule
    C & \textcolor{gray}{\XSolidBrush} &  \textcolor{gray}{\XSolidBrush} & \textcolor{gray}{\XSolidBrush} & \Checkmark & \textcolor{gray}{\XSolidBrush} & 74.36 & \textcolor{gray}{53.82} \\
    D & \textcolor{gray}{\XSolidBrush} & \textcolor{gray}{\XSolidBrush} & \Checkmark &  \Checkmark & \textcolor{gray}{\XSolidBrush} & 63.11 & \textcolor{gray}{23.44} \\
    \midrule
    E & \Checkmark & \Checkmark & \Checkmark & \Checkmark & \textcolor{gray}{\XSolidBrush} & 62.05 & 21.65 \\
    F & \Checkmark & \Checkmark & \Checkmark & \Checkmark & \Checkmark & 60.98 & 15.38 \\
    \bottomrule
\end{tabular}
\vspace{-0.4cm}
\label{tab:ablation}
\end{table} 

\subsection{Analyses and Discussions}

\begin{figure*}[t] 

    \begin{subfigure}[t]{0.24\textwidth}
        \includegraphics[width=\linewidth]{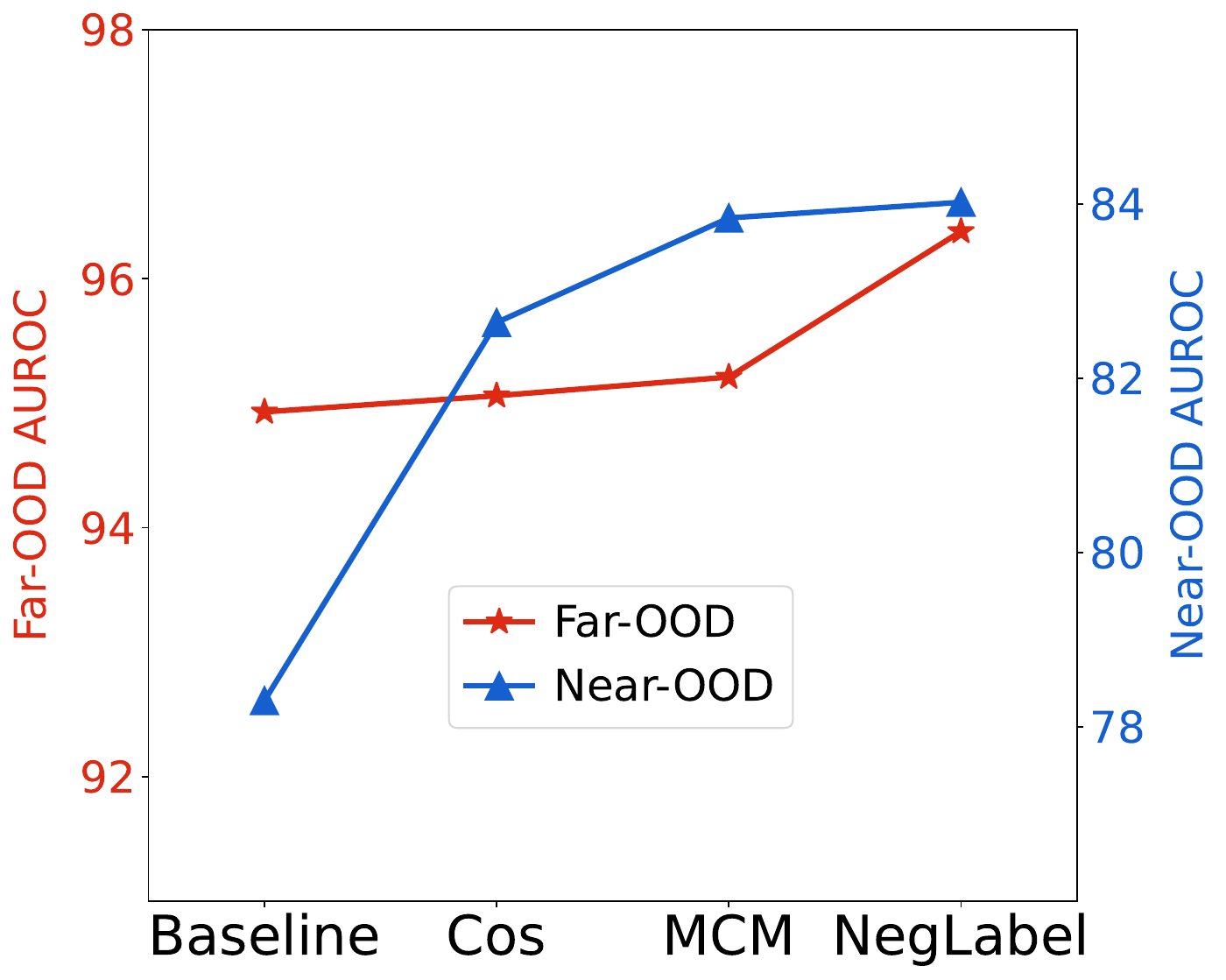}
        \caption{Initial OOD detector}
        \label{fig:init_ood}
    \end{subfigure}
    \hfill 
    \begin{subfigure}[t]{0.24\textwidth}
        \includegraphics[width=\linewidth]{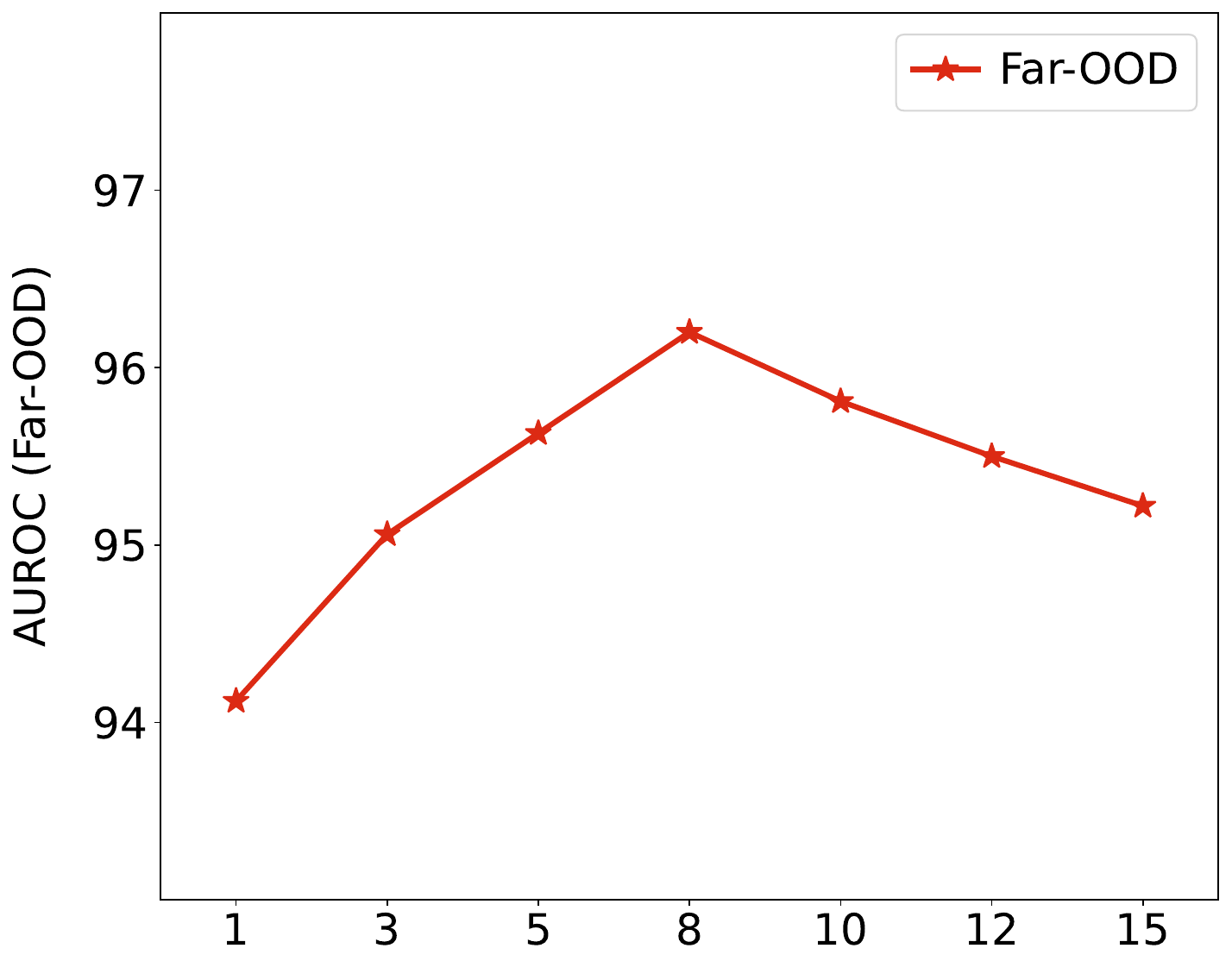}
        \caption{The lengths of negative sentences.}
        \label{fig:lengths}
    \end{subfigure}
    \hfill 
    \begin{subfigure}[t]{0.24\textwidth}
        \includegraphics[width=\linewidth]{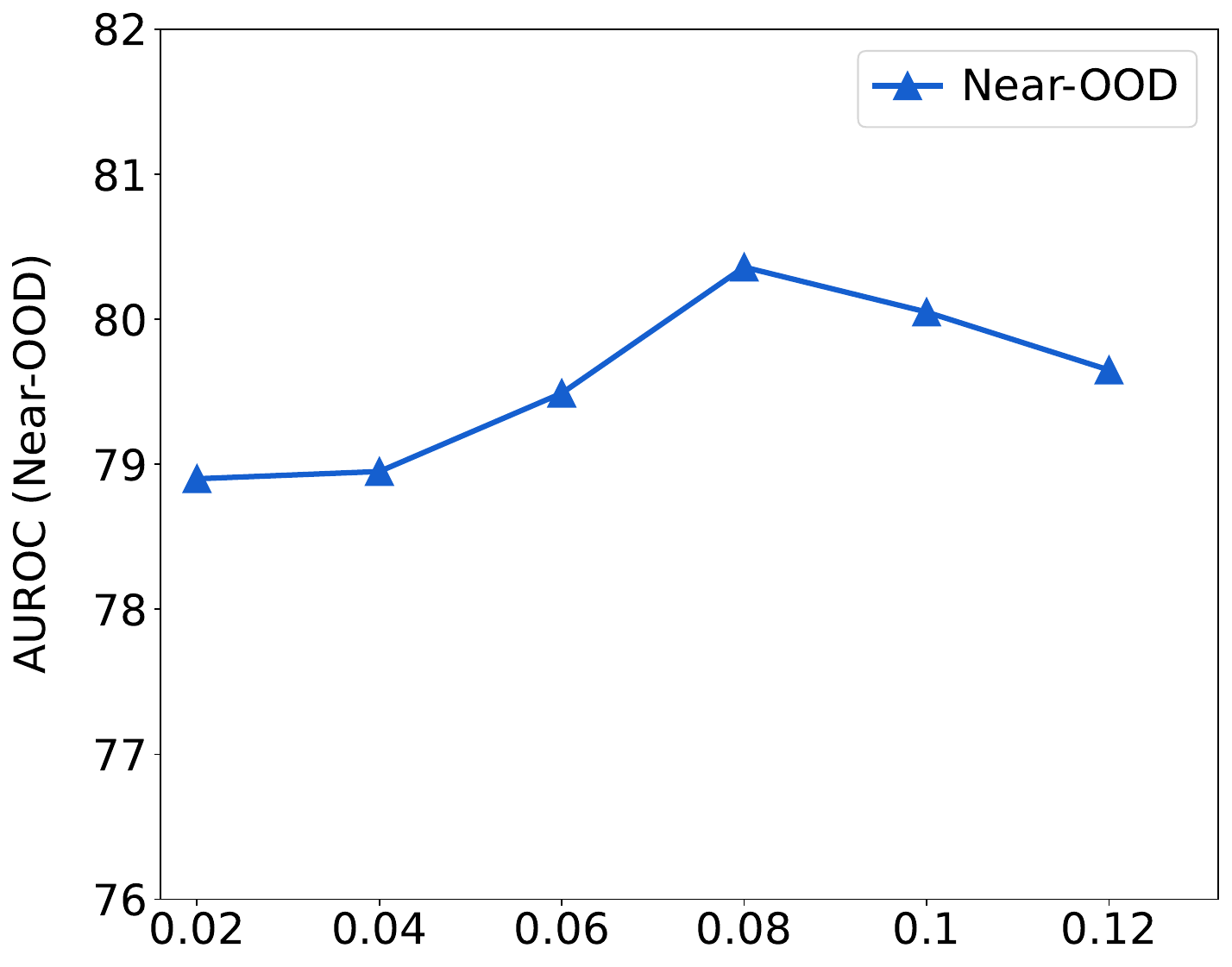}
        \caption{Selection ratio $\delta$}
        \label{fig:subid_delta}
    \end{subfigure}
    \hfill 
    \begin{subfigure}[t]{0.24\textwidth}
        \includegraphics[width=\linewidth]{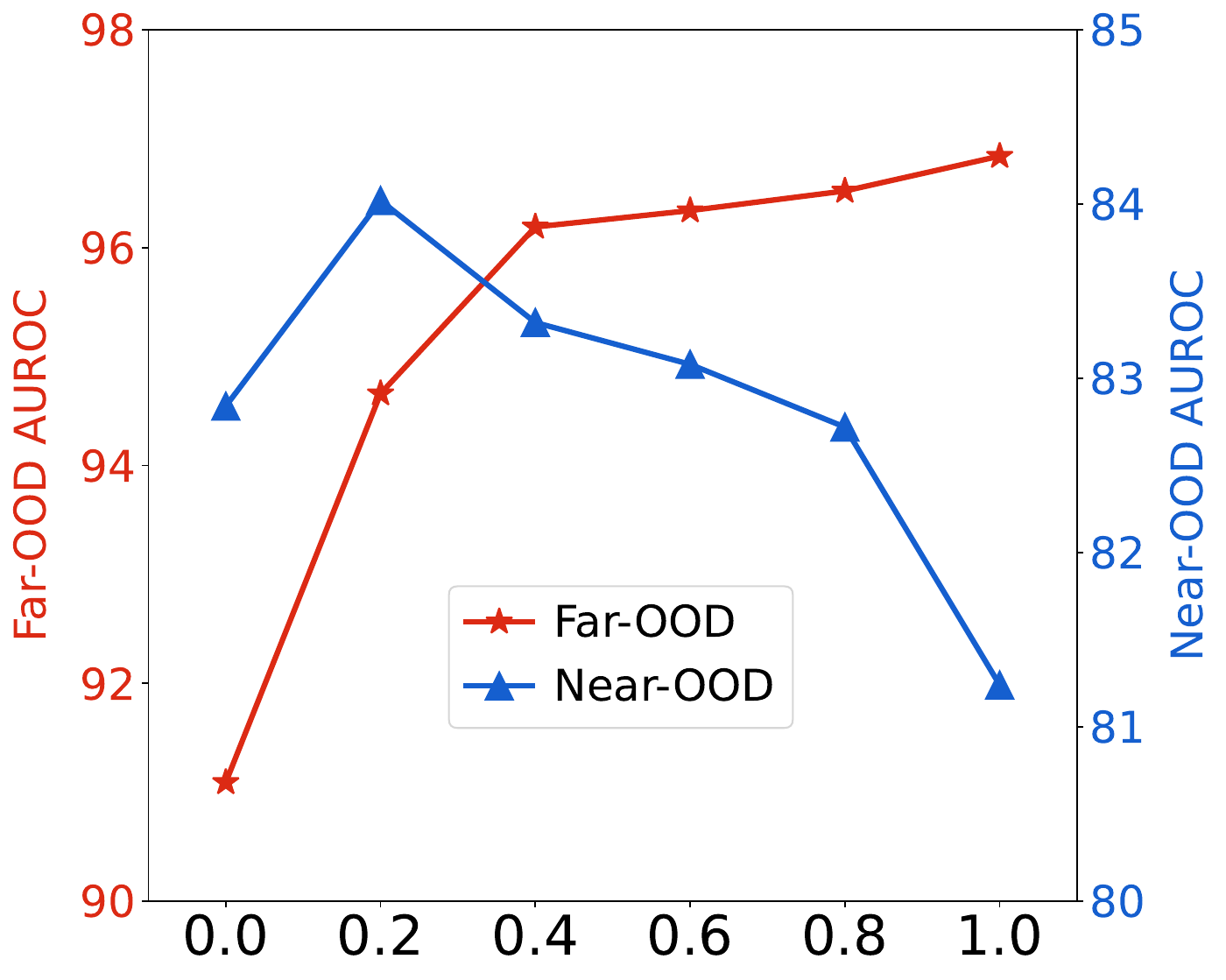}
        \caption{Weight $\lambda$}
        \label{fig:ada_lambda}
    \end{subfigure}
    \hfill 
    \begin{subfigure}[t]{0.24\textwidth}
        \includegraphics[width=\linewidth]{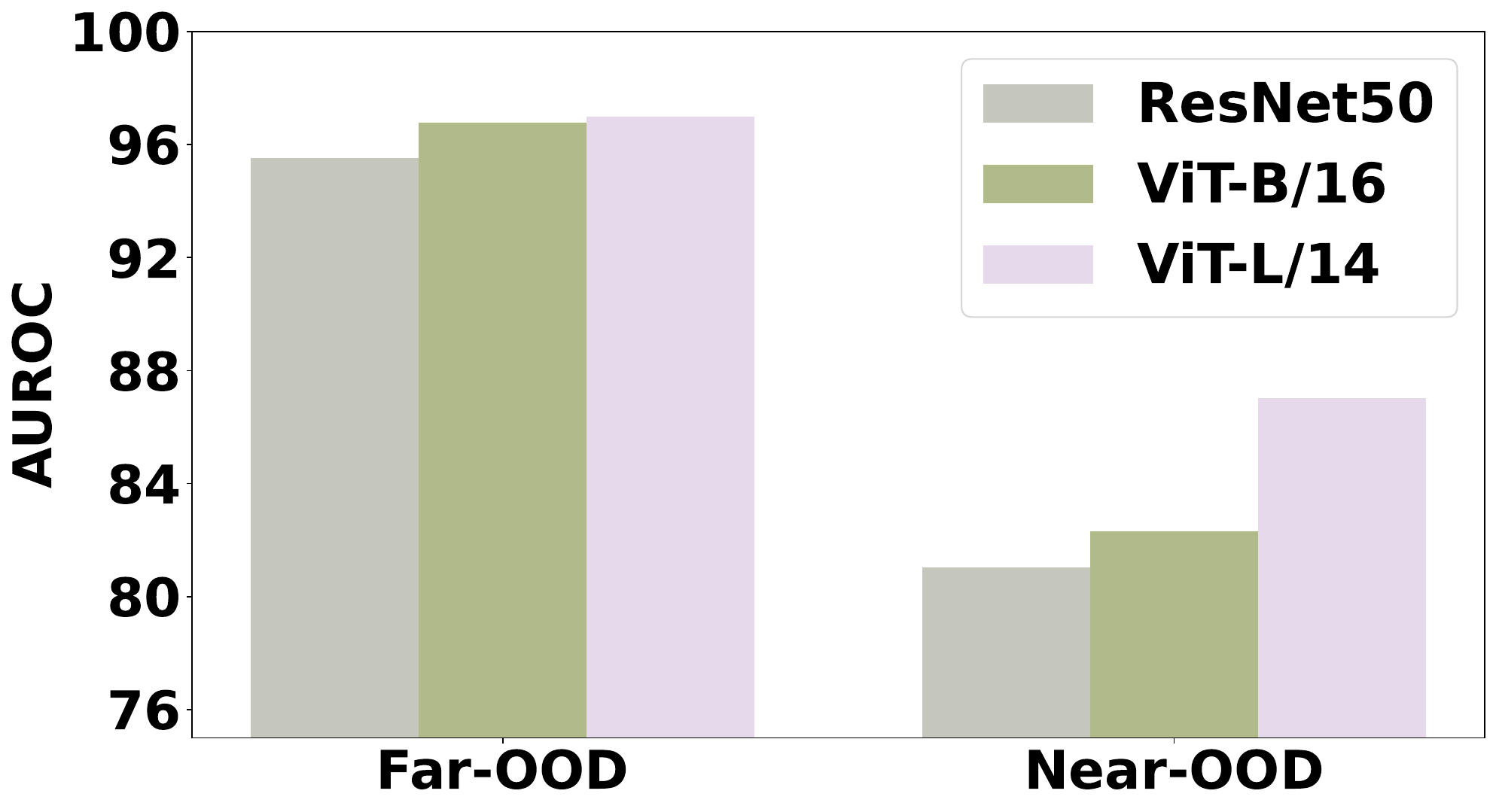}
        \caption{Backbones}
        \label{fig:backbones}
    \end{subfigure}
    \hfill 
    \begin{subfigure}[t]{0.24\textwidth}
        \includegraphics[width=\linewidth]{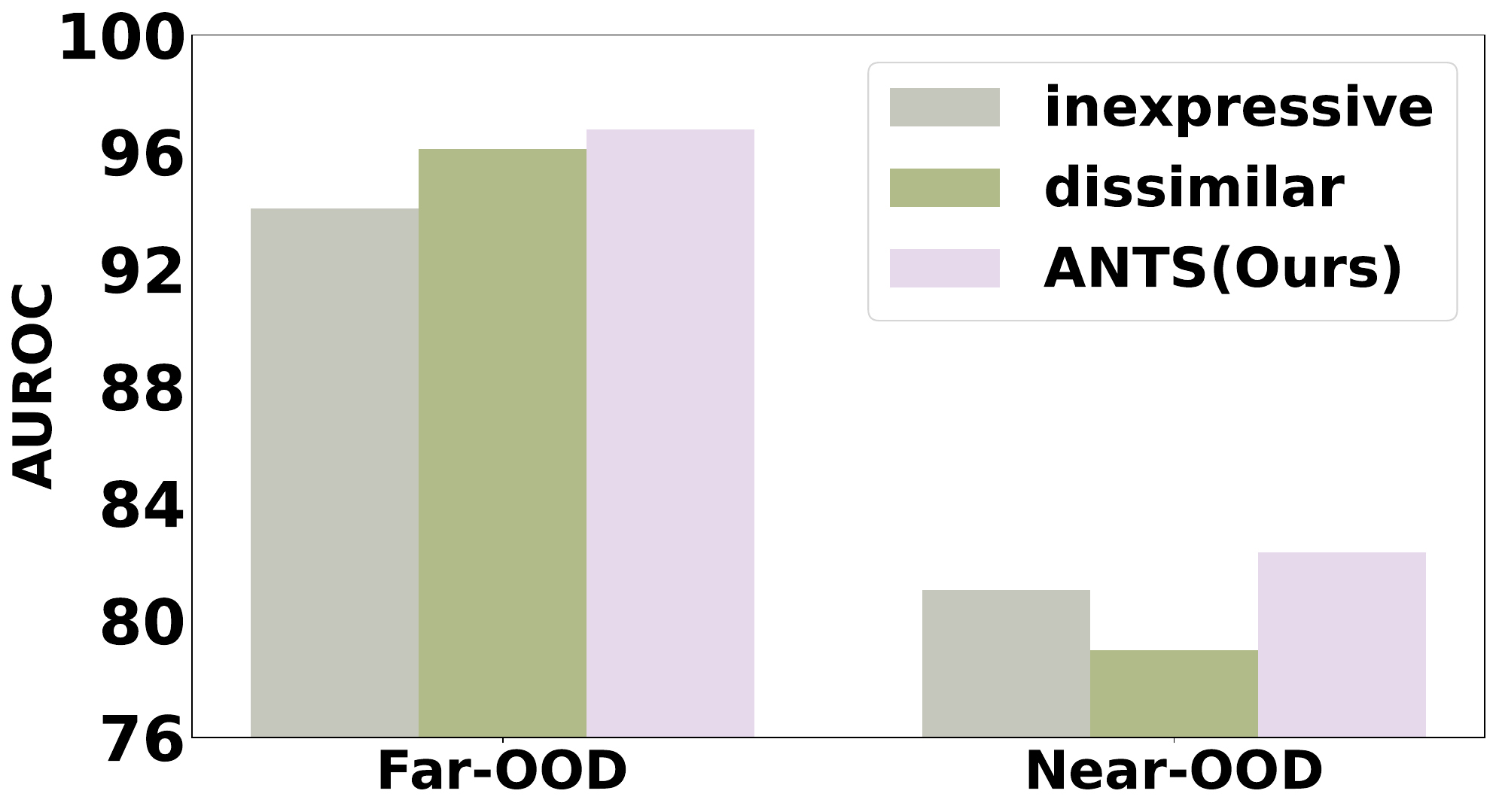}
        \caption{MLLMs prompts}
        \label{fig:prompts}
    \end{subfigure}
    \hfill 
    \begin{subfigure}[t]{0.24\textwidth}
        \includegraphics[width=\linewidth]{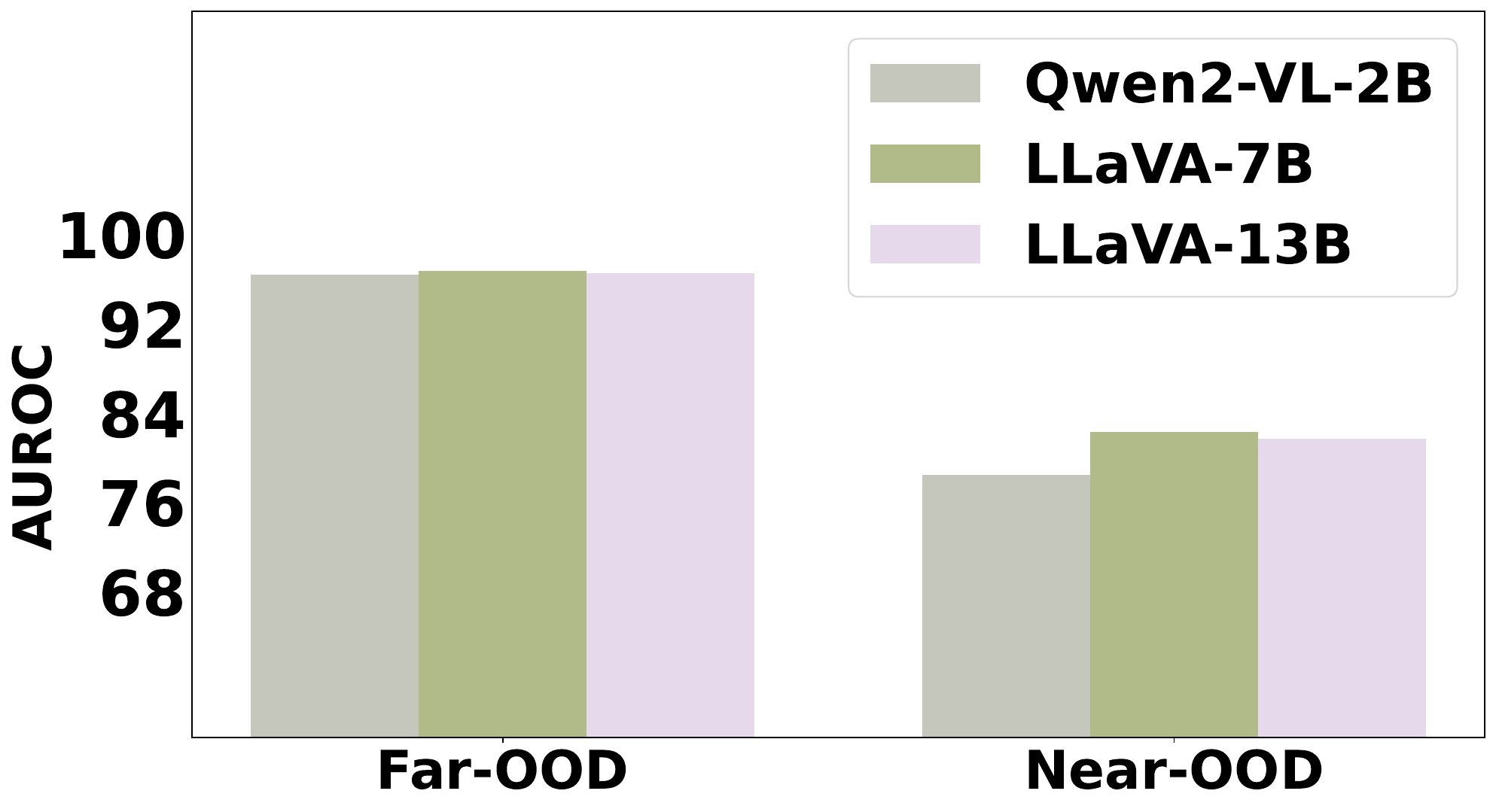}
        \caption{MLLMs}
        \label{fig:mllm}
    \end{subfigure}
    \begin{subfigure}[t]{0.24\textwidth}
        \includegraphics[width=\linewidth]{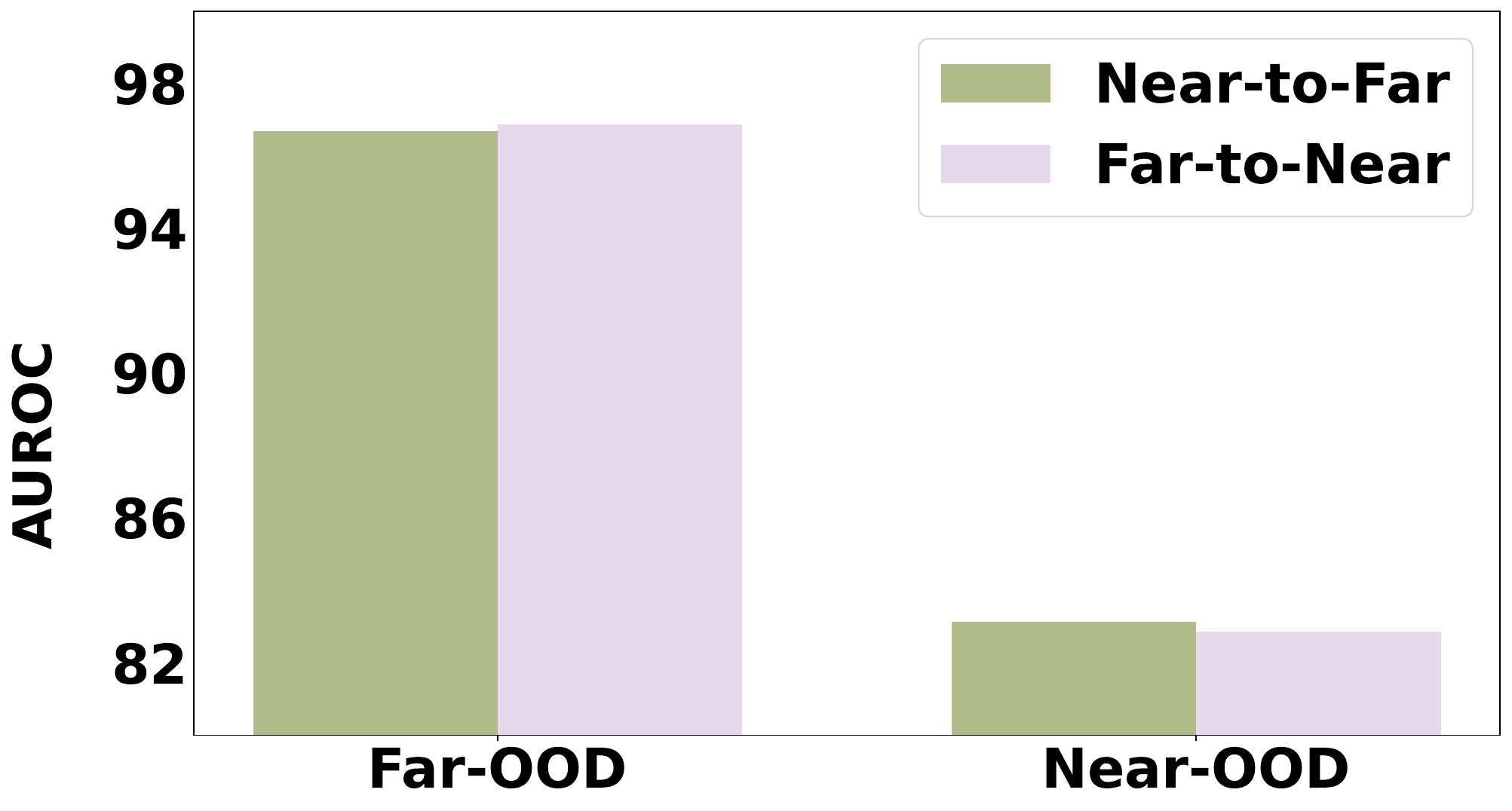}
        \caption{Temporal shift}
        \label{fig:temporal_shift}
    \end{subfigure}
        \vspace{-0.2cm}
    \caption{Analysis on (a) different initial OOD detectors, (b) the lengths of negative sentences, (c) selection ratio $\delta$,  (d) weight $\lambda$, (e) CLIP image encoder backbones, (f) MLLMs prompts, and (g) different MLLMs. (h) Temporal shift. We use Texture~\cite{cimpoi2014describing} and NINCO~\cite{bitterwolf2023or} datasets as Far-OOD and Near-OOD, respectively.}
    \label{fig:analysis}
    \vspace{-0.2cm}
\end{figure*}

\noindent\textbf{Ablation Study.}
As illustrated in Tab.~\ref{tab:ablation}, it is necessary to introduce ENS $\mathcal{Y}^-_{ens}$ with mined negative images, as validated by the advantages of setting B over A in the far-OOD setting. Setting B significantly outperforms NegLabel, confirming the superiority of ENS over NLs. Generating visually similar labels to the mined ID class subset can significantly reduce false negative labels, as justified by the advantages of setting D over C.  
Combining ENS with VSNL by setting $\lambda = 0.5$ balances the results across different OOD sets, as shown in setting E, while using an adaptive $\lambda$ leads to the best results in both OOD scenarios, as shown in setting F.

\noindent\textbf{Analyses on initial OOD detectors.}
Besides NegLabel, we also tested two other variants: (1) a weak MCM detector, and (2) a cosine-distance filter that selects negative  far from ID labels in the feature space. As shown in Fig. ~\ref{fig:init_ood}, even with these weaker detectors, our method still outperforms previous SOTA baseline.

\noindent\textbf{Analyses on the lengths of negative sentences.}
Due to the hallucination issue in MLLMs, we analyzed the length of generated negative sentences. As shown in Fig.~\ref{fig:lengths}, appropriate increases in length enhance expressiveness and OOD detection, while excessive text introduces less discriminative words and hinders performance. Our ENS achieves an optimal balance at an average length of 8.4.

\noindent\textbf{Ratio $\delta$.}
As shown in Fig.~\ref{fig:subid_delta}, generating visually similar labels for all ID classes ( $\delta=1$) performs poorly due to numerous false negative labels, as illustrated in Fig.~\ref{fig:vis_false_negative_labels}. However, using a too small $\delta$ will fail to adequately cover the OOD distribution. We set $\delta=0.08$ in all experiments, although it is not optimal for specific datasets.

\noindent\textbf{Weight $\lambda$.}
As shown in Fig.~\ref{fig:ada_lambda}, a larger $\lambda$ emphasizes the $S_{ens}(x)$ score, improving far-OOD detection, while a smaller $\lambda$ prioritizes the $S_{vsnl}(x)$ score, enhancing near-OOD detection. Our adaptive strategy automatically selects a suitable $\lambda$ for various OOD settings.

\noindent\textbf{Different Backbones.}
As illustrated in Fig.~\ref{fig:backbones}, larger visual backbones generally achieve improved OOD detection. Besides, our ANTS can generalize well to various VLM backbones,  demonstrating its robustness.

\noindent\textbf{Different Prompts for MLLMs.}
We designed two alternative prompts: an inexpressive prompt that limits negative image descriptions to under three words, and a dissimilar prompt that requests negative categories visually distinct from high-frequency ID classes. As shown in Fig.~\ref{fig:prompts}, OOD detection performance declines with both alternatives, confirming our proposed prompts' efficacy.

\noindent\textbf{Various MLLMs.}
When constructing the adaptive negative space, MLLMs of all sizes showed comparable far-OOD detection, but larger models excelled in near-OOD settings. As shown in Fig.~\ref{fig:mllm}, LLaVA-7B performed best, owing to stronger reasoning.

\noindent\textbf{Analysis of Temporal Shift.}
We evaluated ANTS under different temporal shifts.
As shown in Fig.~\ref{fig:temporal_shift}, "Near-to-Far" and "Far-to-Near" indicate testing first on near (or far) OOD, then on the opposite, ANTS maintains strong performance, demonstrating robustness to temporal shift.


\noindent\textbf{Complexity Analyses.}
As analyzed in Tab.~\ref{tab:module_timing} and Tab.~\ref{tab:complexity}, ANTS requires no learnable parameters. Although individual MLLM calls (ENS/VSNL) have higher latency, they are only selectively triggered for a small subset of samples. By amortizing these costs and utilizing a compact MLLM, ANTS maintains a competitive inference speed of 2.84 ms/image, as most samples are processed solely by the CLIP encoder.



\begin{table}[h]
\vspace{-0.3cm}
\small
\centering
\caption{Latency (ms) breakdown (ImageNet).}
\begin{tabular}{lcccc}
\toprule
Modules & ENS & VSNL & CLIP & Others \\ \midrule
Per-call Latency  & 72.55 & 28.75 & 2.59 & 0.02 \\
Avg. Latency/Img  & 0.18 & 0.06 & 2.59 & 0.02 \\ \bottomrule
\end{tabular}
\label{tab:module_timing}
\end{table}

\begin{table}[h]
\centering
\caption{Complexity analyses. All results are obtained by using a GeForce RTX 3090 GPU.}
\small
\label{tab:complexity} 
\begin{tabular}{l|ccc}
\hline 
Methods & Train Time & Latency (ms) & Param. (M) \\
\hline 
ZOC & $>24 \mathrm{~h}$ & 5.38 & 336 \\
CLIPN & $>24 \mathrm{~h}$ & 2.53 & 37.8 \\
\hline 
EOE & - & 2.78 & - \\
NegLabel & - & 2.61 & - \\
AdaNeg & - & 2.70 & - \\
\textbf{ANTS} & - & \textbf{2.84} & - \\
\hline
\end{tabular}
\end{table}

\section{Conclusion and Future Work}
This paper presents ANTS, a training-free, zero-shot framework for out-of-OOD detection. We first investigate three limitations of existing NLs methods. To address these issues, ANTS caches negative images and visually similar ID classes from historical test images, leveraging test-time MLLM understanding and reasoning through tailored prompts to construct a more accurate adaptive negative textual space. Two noise-filtering strategies are introduced to mitigate interference from ID noise and false negative labels. Finally, an adaptive scoring mechanism dynamically balances the two textual spaces, enhancing the framework's scalability across diverse OOD scenarios. Experimental results demonstrate that ANTS achieves state-of-the-art performance on zero-shot OOD detection benchmarks.

One minor limitation of our approach is that utilizing the MLLM model during testing necessitates GPU memory. More efficient utilization of MLLMs during the testing phase presents a meaningful direction for future work.




\clearpage
{
    \small
    \bibliographystyle{ieeenat_fullname} 
    \bibliography{main}
}


\end{document}